\documentclass[twoside]{article}

\usepackage[utf8]{inputenc} 
\usepackage[T1]{fontenc}    
\usepackage[hidelinks]{hyperref}       
\usepackage{url}            
\usepackage{booktabs}       
\usepackage{amsfonts}       
\usepackage{nicefrac}       
\usepackage{microtype}      
\usepackage{wrapfig}
\usepackage{graphicx}
\usepackage{amsmath,amssymb,amsthm}
\usepackage{verbatim}
\usepackage{algpseudocode}
\usepackage{algorithm}
\usepackage{algorithmicx}
\usepackage{bm}
\usepackage{csquotes}
\usepackage{xfrac}

\newcommand{\pred}{\mathrm{pred}}
\newcommand{\prob}{\mathbb{P}}
\newcommand{\expect}{\mathbb{E}}

\newcommand{\Log}{\mathrm{Log}}
\newcommand{\Exp}{\mathrm{Exp}}
\newcommand{\Riem}{\mathrm{Riem}}
\newcommand{\Tr}{\mathrm{Tr}}

\newcommand{\R}{\mathbb{R}}
\newcommand{\N}{\mathcal{N}}
\newcommand{\GP}{\mathcal{GP}}

\newcommand{\Cov}{\text{Cov}}
\newcommand{\vect}[1]{\bm{#1}}

\theoremstyle{plain}

\usepackage{graphicx}
\usepackage{color}
\usepackage{pstricks}

\usepackage{url}  
\usepackage{cite}

\usepackage[accepted]{aistats2019}
\usepackage[round]{natbib}

\begin{document}
\setlength{\abovedisplayskip}{5pt}
\setlength{\belowdisplayskip}{5pt}
%

%
\runningauthor{Mallasto, Hauberg, Feragen}

\twocolumn[

\aistatstitle{Probabilistic Riemannian submanifold learning with wrapped Gaussian process latent variable models}

\aistatsauthor{Anton Mallasto \And S{\o}ren Hauberg  \And  Aasa Feragen }

\aistatsaddress{  Department of Computer Science \\ University of Copenhagen  \And  DTU Compute \\ Technical University of Denmark \And Department of Computer Science \\ University of Copenhagen  } 
]

\begin{abstract}
Latent variable models (LVMs) learn probabilistic models of data manifolds lying in an \emph{ambient} Euclidean space. In a number of applications, a priori known spatial constraints can shrink the ambient space into a considerably smaller manifold.  Additionally, in these 
applications the Euclidean geometry might induce a suboptimal similarity measure, which could be improved by choosing a different metric. Euclidean models ignore such information and assign probability mass to data points that can never appear as data, and vastly different 
likelihoods to points that are similar under the desired metric. We propose the wrapped Gaussian process latent variable model (WGPLVM), that extends Gaussian process latent variable models to take values strictly on a given ambient Riemannian manifold, making the model blind to impossible data points. This allows non-linear, probabilistic inference of low-dimensional Riemannian submanifolds from data. Our evaluation on diverse datasets show that we improve performance on several tasks, including encoding, visualization and uncertainty quantification.
\end{abstract}
\section{INTRODUCTION}
Unsupervised learning aims at modelling structure in unlabeled data, such as its geometry. Sometimes, information on this geometry is available through spatial constraints or a non-Euclidean metric, e.g. the data lives on a Riemannian manifold. Incorporating the known Riemannian manifold in a probabilistic model should improve model fit, and save us from learning what we already know. In this work, we study a probabilistic latent variable model that takes the geometry into account.

\begin{figure}
\centering
\includegraphics[width = 0.35\textwidth]{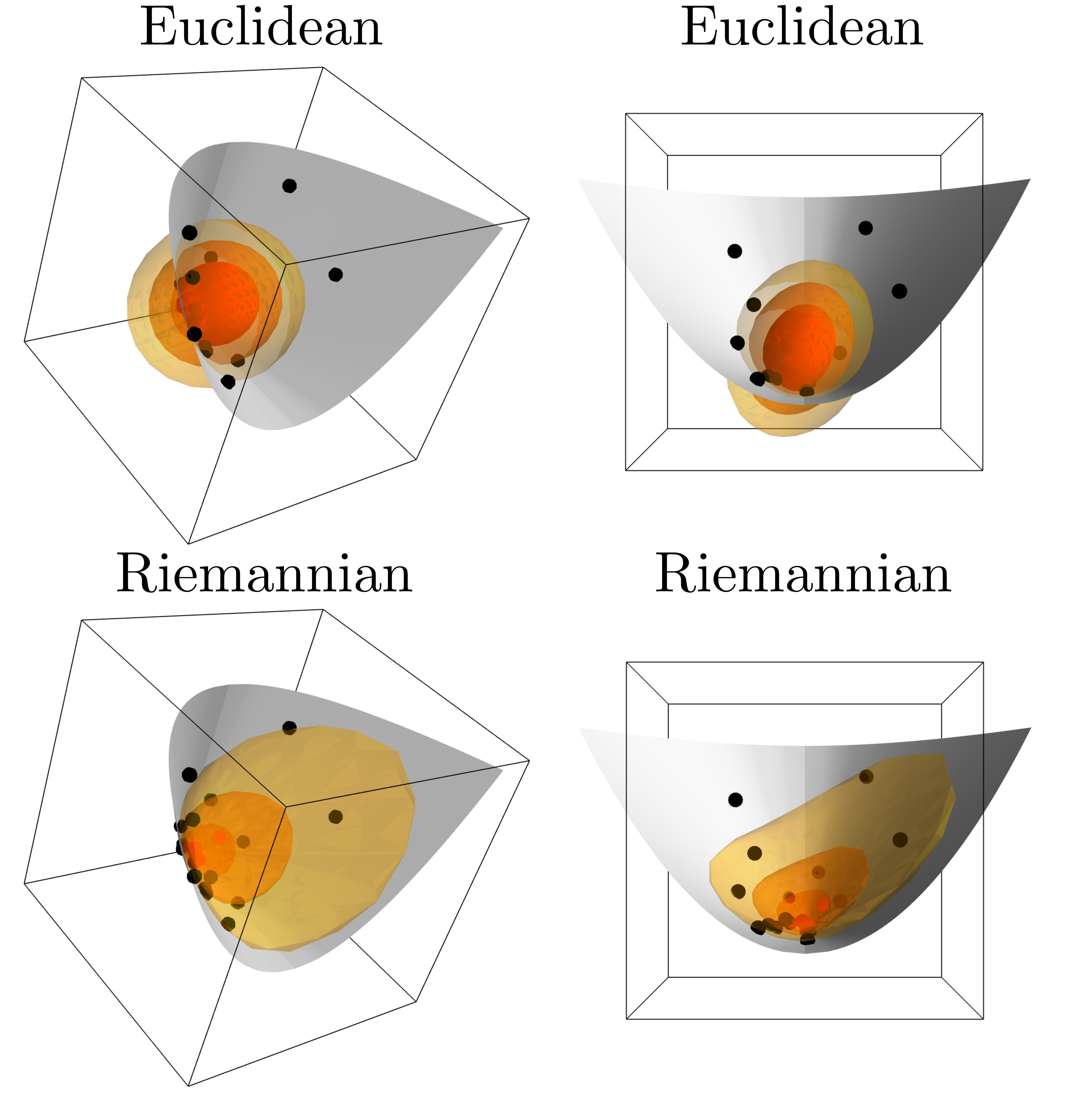}
\caption{The ambient manifold $SPD(2)$ is the open subset \emph{on the inside} of the visualized grey cone in the ambient Euclidean space $\R^3$. \textbf{Top row:} A Euclidean Gaussian distribution fitted to a set of $SPD(2)$ matrices (black dots) escapes \emph{outside} of $SPD(2)$. \textbf{Bottom row:} The Riemannian Log-Euclidean metric yields a wrapped Gaussian distribution that remains inside $SPD(2)$, providing a better fit to the data. The colored trust regions are confidence regions of the (W)GDs.} 
\label{fig:spd2}
\vspace{-0.2in}
\end{figure}

\textbf{Where do manifolds come from?} Data points on a sphere are forced to have norm one, covariance matrices are symmetric and positive definite, and shapes do not depend on scale, rotation or placement. Enforcing such constraints or invariances, one replaces the ambient Euclidean space by an ambient \emph{manifold}. The \emph{ambient space} refers to the set of all those points, which the model views as possible data points. The constraints alter the shortest paths between data objects, giving rise to a \emph{Riemannian metric}. Riemannian metrics can also be imposed by modelling choices; closeness under the Euclidean metric does not always express desired similarity of data objects. These metrics can be learned from data \citep{hauberg2012} or imposed based on domain knowledge \citep{arsigny06}.

\textbf{Euclidean probabilistic models on manifold data} assign probability mass to impossible data points under spatial constraints. Furthermore, points that are similar under the chosen non-Euclidean metric can be assigned very different likelihoods, which can cause a poor fit to the data. Both issues affect especially the uncertainty estimates. These issues can be avoided by exploiting the Riemannian geometry in the model. Fig.~\ref{fig:spd2} shows points in $SPD(2)$, the space of $2\times 2$ symmetric positive-definite matrices, with fitted Euclidean and Riemannian models. The points outside the cone are not  $SPD(2)$ matrices. Under the Log-Euclidean metric, which generalizes the log transform to matrices, elements on the boundary (in gray) lie infinitely far from interior points. The metrics, and hence the induced models, are vastly different. This results in the Riemannian model with an improved model fit.

\textbf{Contributions.} Motivated by these observations, we introduce the \emph{wrapped Gaussian process latent variable model} (WGPLVM). This extends the Gaussian process latent variable model (GPLVM) to data on Riemannian manifolds by employing \emph{wrapped Gaussian processes} (WGPs). Like the GPLVM, the WGPLVM defines a probabilistic model between elements in a lower dimensional \emph{latent space} and the data, providing uncertainty estimates. As WGPs take values strictly on a given Riemannian manifold, the WGPLVM enforces known constraints and invariances, and accounts for modelling choices concerning the metric. 

We demonstrate the WGPLVM on several different manifolds and tasks. We show that our method provides more efficient encoding of the original data compared to the Euclidean GPLVM, provides superior uncertainty estimates and better captures trends in the data, resulting in improved visualization results.

\textbf{Related Literature.} First, we discuss methods in \emph{manifold learning}, which view data points as elements of a Euclidean space. Then, we discuss related work in \emph{submanifold learning}, that works strictly on Riemannian manifolds. Note that some manifold learning methods can impose known geometry on the latent space. Models relying on kernels (e.g. the GPLVM and WGPLVM) can encode such structure on the latent space \citep{lin17}. This is different from imposing geometric constraints on the data space.

\emph{Manifold learning} infers a low-dimensional manifold that captures the trend of given data. Classical algorithms~\citep{isomap, lle, lapeigmaps} learn a low distortion projection from a data submanifold of the original, Euclidean ambient space, onto a low-dimensional Euclidean space. Latent variable models (LVMs)~\citep{autoencoders,gans,gplvm} learn the reverse \emph{latent embedding} from the latent space into the ambient space, associating each point in the latent space with an ambient space point. In the well-known \emph{Gaussian process latent variable model} (GPLVM)~\citep{gplvm}, the latent embedding is a Gaussian process (GP) over the latent space, and hence learns not only a manifold embedding into $\R^n$, but also a model of its uncertainty. GPLVMs have inspired other LVMs \citep{titsias10, lawrence07, urtasun07}, that all rely on Euclidean geometry. \citet{urtasun08} consider topologically constrained LVMs and \citet{varol12} consider GPLVMs with spatial constraints, where the constraints are enforced through slack variables and local linearization. Our method works intrinsically on the specific Riemannian manifold, taking the topology, spatial constraints and the Riemannian metric into account. Thus the WGPLVM falls into the category of submanifold learning.

\begin{figure}
\centering
\includegraphics[width = 0.32\textwidth]{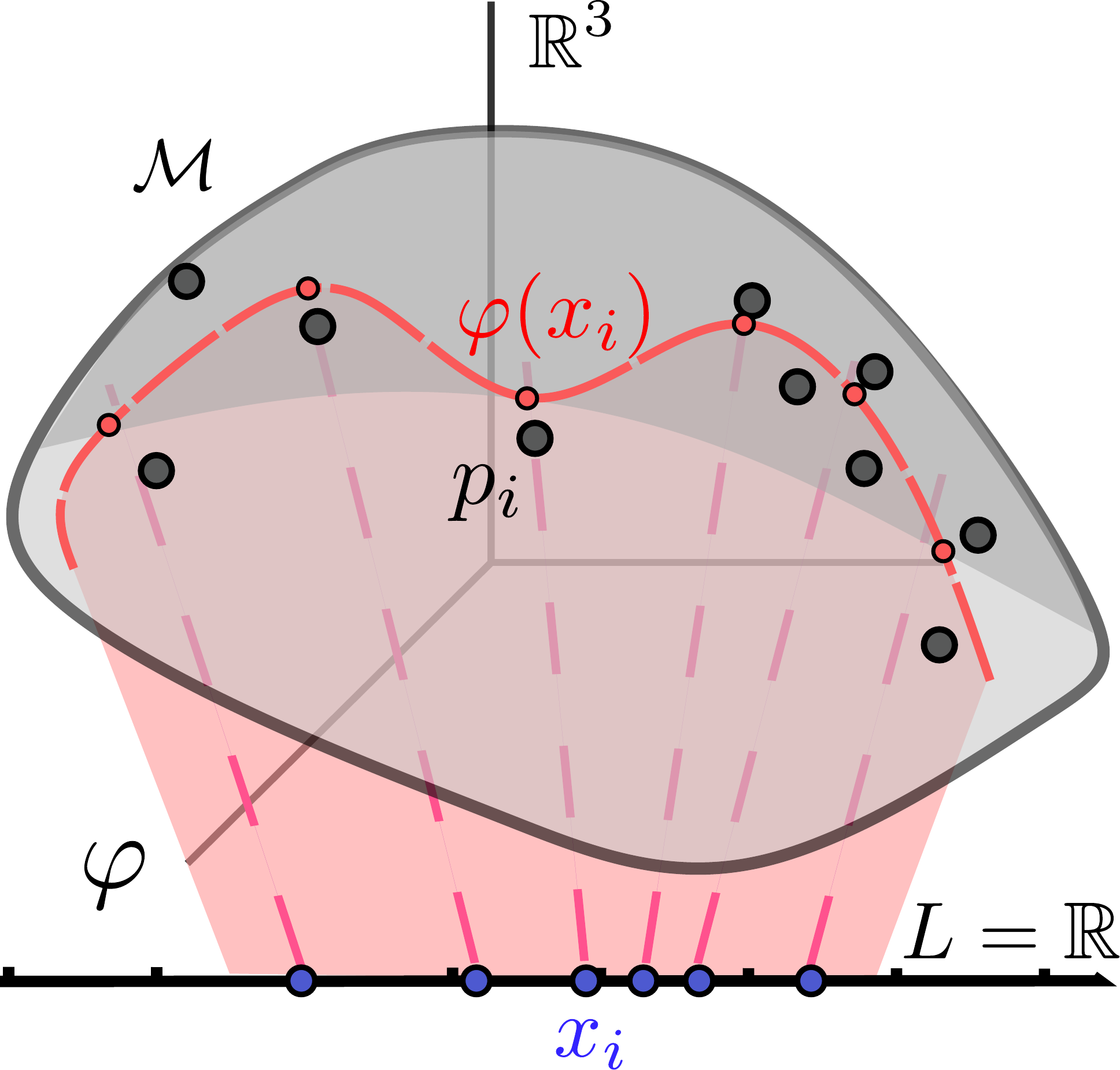}
\caption{Illustration of submanifold learning.} 
\label{LVMIllustration}
\vspace{-0.2in}
\end{figure}

\emph{Submanifold learning} algorithms, illustrated in Fig.~\ref{LVMIllustration}, aim to infer a model $\varphi$ from a latent space $L$ to a submanifold $M$ (dashed red) of a known \emph{ambient manifold} $\mathcal{M}$ of points that satisfy the constraints. The map $\varphi$ associates the data $p_i \in \mathcal{M}$ (dark grey) with latent variables $x_i\in L$ (blue). \emph{Principal geodesic analysis} (PGA)~\citep{fletcher04, huckemann10} estimates geodesic submanifolds, \emph{Riemannian principal curves}~\citep{principalcurves} and \emph{barycentric subspaces}~\citep{barycentric} estimate less constrained submanifolds. \emph{Probabilistic PGA}~\citep{ppga} introduces uncertainty by estimating probabilistic geodesic subspaces. The WGPLVM contributes non-geodesic, probabilistic learning of the submanifold from a prior model, allowing considerable flexibility compared to previous models.

Examples of manifold valued data include directional statistics, which consider spherical data \citep{urtasun20063d,mardia2009directional}, covariance matrices as data objects in economics and computer vision~\citep{GeneralizedWishartProcesses,Tuzel:ECCV:2006} and in diffusion MRI or materials science \citep{batchelor,fletcherjoshi2004}, and statistics of shape, which is of fundamental interest in computer vision \citep{kendall, Freifeld:ECCV:2012}. In each example, the common approach is to incorporate the Riemannian structure in the statistical analysis.

\section{PRELIMINARIES}
This section introduces the necessary preliminaries and notation. We first review Gaussian processes (GPs) and the Gaussian process latent variable model (GPLVM) \citep{lawrence04}. Next, we summarize the necessary concepts from Riemannian geometry. Subsequently, we review the wrapped Gaussian processes (WGPs) introduced by \citet{mallasto:cvpr:2018}, which form the cornerstone of the present work. \label{prel}

\textbf{Gaussian processes}. Let $\N(\mu,\Sigma)$ denote a multivariate Gaussian distribution (GD) with mean $\mu\in \mathbb{R}^d$ and covariance matrix $\Sigma\in \mathbb{R}^{d\times d}$, and write the associated probability density function as $\N(v|\mu,\Sigma)$ for $v\in \mathbb{R}^d$. A \emph{Gaussian process} (GP)  is a collection $f$ of random variables, so that any finite subcollection $(f(\omega_i))_{i=1}^N$ is jointly Gaussian, where $\omega_i\in \Omega$ are elements of the \emph{index set}. Any GP $f$ is uniquely characterized by
\begin{equation}
\begin{aligned}
m(\omega)&=\expect\left[f(\omega)\right],\\
 k(\omega,\omega')&=\expect\left[(f(\omega)-m(\omega))(f(\omega')-m(\omega'))^T\right],
\end{aligned}
\end{equation}
called the \emph{mean function} $m$ and \emph{covariance function} $k$, denoted $f\sim \GP(m,k)$. For more about GPs and their applications, see \citet{rasmussen}.

\textbf{Gaussian process latent variable model.} The Gaussian process latent variable model (GPLVM) is a GP-based dimensionality reduction technique, which aims to learn a probabilistic model relating elements in the low dimensional \emph{latent space} $L\subseteq \mathbb \R^{n'}$ to observed data $Y=\{y_i\}_{i=1}^N\subset \R^n$, with $n' < n$. The model approximates the manifold that $Y$ lives on. The probabilistic model is computed by choosing a prior GP $f\sim \GP(m,k_{\theta})$ with hyper-parameters $\theta \in \Theta$. The hyper-parameters are optimized with the \emph{latent variables} $X=\{x_i\}_{i=1}^N \in L$ to maximize the log-likelihood 
\begin{equation}
\begin{aligned}
\log(\prob(Y|X,\theta)) = & -\frac{nN}{2}\ln(2\pi) - \frac{n}{2}\ln |K_{X,\theta}|\\
  &- \frac{1}{2}\Tr\left(K^{-1}_{X,\theta} YY^T\right),
\end{aligned}
\end{equation}
where $(K_{X,\theta})_{ij} =k_\theta(x_i,x_j)$, and $X,Y$ denote the corresponding data matrices. Finally, we condition the optimal prior $f$ on the chosen latent variables $X$ and data $Y$, to yield the predictive distribution of the model. Note that any prediction $f(x)$ has support in the whole $\R^n$, thus ignoring any constraints or invariances.

In differential geometric terms, a GPLVM can be viewed to learn a stochastic \emph{chart} for the approximate manifold on which the dataset $Y$ lives.

\textbf{Riemannian geometry.} A \emph{Riemannian manifold} is a \emph{smooth manifold} $M$ with a \emph{Riemannian metric}, i.e.~a smoothly varying inner product $g_p(\cdot,\cdot)$ on the tangent space $T_pM$ at each $p\in M$, which induces a distance function $d_M$ on $M$. Each $(p,v)$ in the \emph{tangent bundle} $TM = \bigcup_{p\in M} \left(\{p\}\times T_pM\right)$ defines a \emph{geodesic} $\gamma$ (locally shortest path) on $M$, so that $\gamma(0)=p$ and $\dot{\gamma}(0)=v$. 
 
The Riemannian \emph{exponential map} $\Exp\colon TM\rightarrow M$ is given by $(p,v)\mapsto \Exp_p(v)=\gamma(1)$, where $\gamma$ is the geodesic corresponding to $(p,v)$. The exponential $\Exp_p$ at $p$ is a diffeomorphism between a neighborhood $0\in U_p\subset T_pM$ and a neighborhood $p\in V_p\subset M$, which is chosen in a maximal way to preserve injectivity. The \emph{logarithmic map} $\Log_p\colon V_p\rightarrow T_pM$ is characterized by the identity $\Exp_p(\Log_p(p'))=p'$. Outside of $V_p$, we use $\Log_p(p')$ to denote $v\in\Exp_p^{-1}(p')$ with a minimal norm, chosen in a \emph{measurable} way. The complement of $V_p$ in $M$ is called the \emph{cut-locus} at $p$, where unique geodesics cannot be defined. Multiple useful manifolds have empty cut-locus, so that $V_p = M$, including manifolds with non-positive curvature as well as the space of positive-definite symmetric matrices used below.

Let $\Exp_p(v) = q$ and $\gamma(t) = \Exp_p(tv)$. The differential $D_p\Log_p(q)$ (in some coordinate chart) is given by (see supplementary material for \citep{pennec16barycentric}) 
\begin{equation}
D_p\Log_p(q) = \left(J_0(1)\right)^{-1}J_1(1),
\label{logcovder}
\end{equation}
where $J_i$ are Jacobi fields solving the linear ordinary differential equation
\begin{equation}
\ddot{J}_i(t) + R(t)J_i(t) = 0,
\end{equation}
with initial conditions $J_0(0) = 0$, $\dot{J}_0(0) = I_n$, and $J_1(0) = I_n$, $\dot{J}_1(0) = 0$. Here $R(t)$ is given by $R_{ij} = \langle \Riem_{\gamma(t)}(\dot{\gamma}(t), e_i(t))\dot{\gamma}(t), e_j(t)\rangle_{\gamma(t)}$ and $(e_1(t),...,e_n(t))$ is an orthonormal basis for $T_{\gamma(t)}M$, defined by $e_1(0) = \frac{v}{\|v\|_2}$ and each $e_j(t)$ evolves through parallel transportation. Furthermore, $\Riem_{\gamma{t}}$ denotes the curvature tensor and $I_n$ is the $n$-by-$n$ identity matrix, where $n$ is the dimension of the manifold. For a thorough exposition in Riemannian geometry, see \citep{docarmo}.

Let $M_i$ be Riemannian manifolds with metrics $g_i$, exponential maps $\Exp^i$ and logarithmic maps $\Log^i$ for $i=1,2$. Then $M=M_1\times M_2$ turns into a Riemannian manifold when endowed with the metric $g=g_1+g_2$, which has the component-wise computed exponential map $\Exp_{(p_1,p_2)}((v_1,v_2))=\left(\Exp^1_{p_1}(v_1),\Exp^2_{p_2}(v_2)\right)$. The logarithmic map $\Log$ on the product manifold is defined likewise.

\begin{figure}
\centering
\includegraphics[width = 0.25\textwidth]{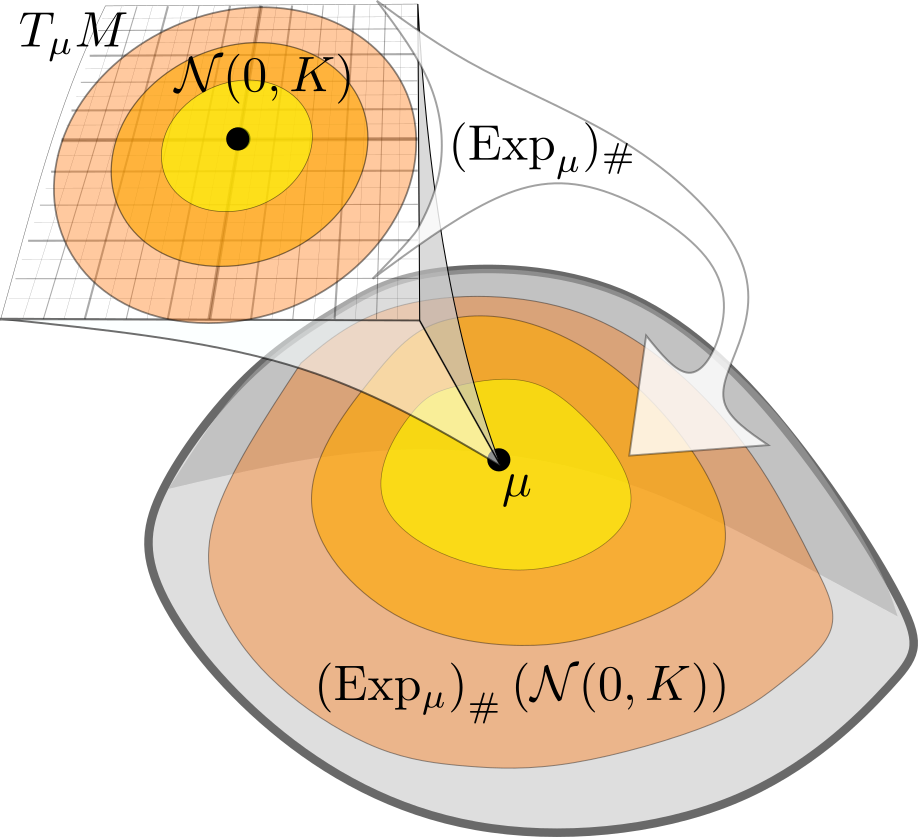}
\caption{WGDs defined as a Gaussian $\N(0,K)$ in the tangent space $T_\mu M$ over the basepoint $\mu$, which is pushed forward by the exponential map $\Exp_\mu$ to $M$.} 
\label{WGDIllustration}
\vspace{-0.1in}
\end{figure}
\textbf{Wrapped Gaussian distributions.} Let $(M,g)$ be an $n$-dimensional geodesically complete Riemannian manifold. Let $\nu$ be a measure on $X$ and $f\colon X\rightarrow Y$ be a measurable map. We define the \emph{push-forward} as $f_\#\nu(A) := \nu(f^{-1}(A))$ for any measurable set $A$ in $Y$. A random point $X$ on $M$ follows a \emph{wrapped Gaussian distribution} (WGD), if for some $\mu\in M$ and a symmetric positive definite matrix $K\in \mathbb{R}^{n\times n}$ 
\begin{equation}
X\sim \left(\Exp_\mu\right)_\#\left(\N(0,K)\right),
\end{equation}
denoted by $X\sim \N_M(\mu,K)$. The WGD is thus defined by a GD $\N(0,K)$ in the tangent space $T_\mu M$, that is pushed-forward onto $M$ by the exponential map $\Exp_\mu$ (see Fig.~\ref{WGDIllustration}). We call $\mu=:\mu_{\N_M}(X)$ the \emph{basepoint} of $X$, and $K=: \Cov_{\N_M}(X)$ the \emph{tangent space covariance}. 

Two random points $X_i\sim \N_{M_i}(\mu_i,K_i)$, $i=1,2$ are \emph{jointly WGD}, if $(X_1,X_2)$ is a WGD on the product manifold $M_1\times M_2$, given by
\begin{equation}
(X_1,X_2)
\sim \N_{M_1\times M_2}\left(\begin{pmatrix}
\mu_1\\
\mu_2
\end{pmatrix},\begin{pmatrix}
K_1&K_{12}\\
K_{21}&K_2
\end{pmatrix}\right),
\label{jntdist}
\end{equation}
for some matrix $K_{12}=K_{21}^T$. Then, $X_1$ can be conditioned on $X_2$, resulting in a push-forward of a Gaussian mixture in $T_{\mu_1}M_1$ by the exponential map
\begin{equation}
X_1|(X_2=p_2) \sim  \left(\Exp_{\mu_1}\right)_\#\left(\sum_{v\in A}\lambda_v\N\left(\mu_v,K_v\right)\right),
\label{cd_wgd}
\end{equation}
where $A = \{v\in T_{\mu_2}M~|~\Exp_{\mu_2}(v)=p_2\}$ is the preimage of $p_2$.
The means and covariance matrices of the Gaussian mixture components are given by
\begin{equation}
\mu_v = K_{12}K_2^{-1}v,\qquad K_v = K_1-K_{12}K_2^{-1}K_{12}^T,
\end{equation}
and the component weights are
\begin{equation}
 \lambda_v= \frac{\N(v|\vect{0},K_2)}{\prob\{A\}},~~\prob\{A\}= \sum_{v\in A} \N(v|\vect{0},K_2).
\label{cd_wgd2}
\end{equation}

\textbf{Wrapped Gaussian processes.}
\begin{figure}
\centering
\includegraphics[width = 0.25\textwidth]{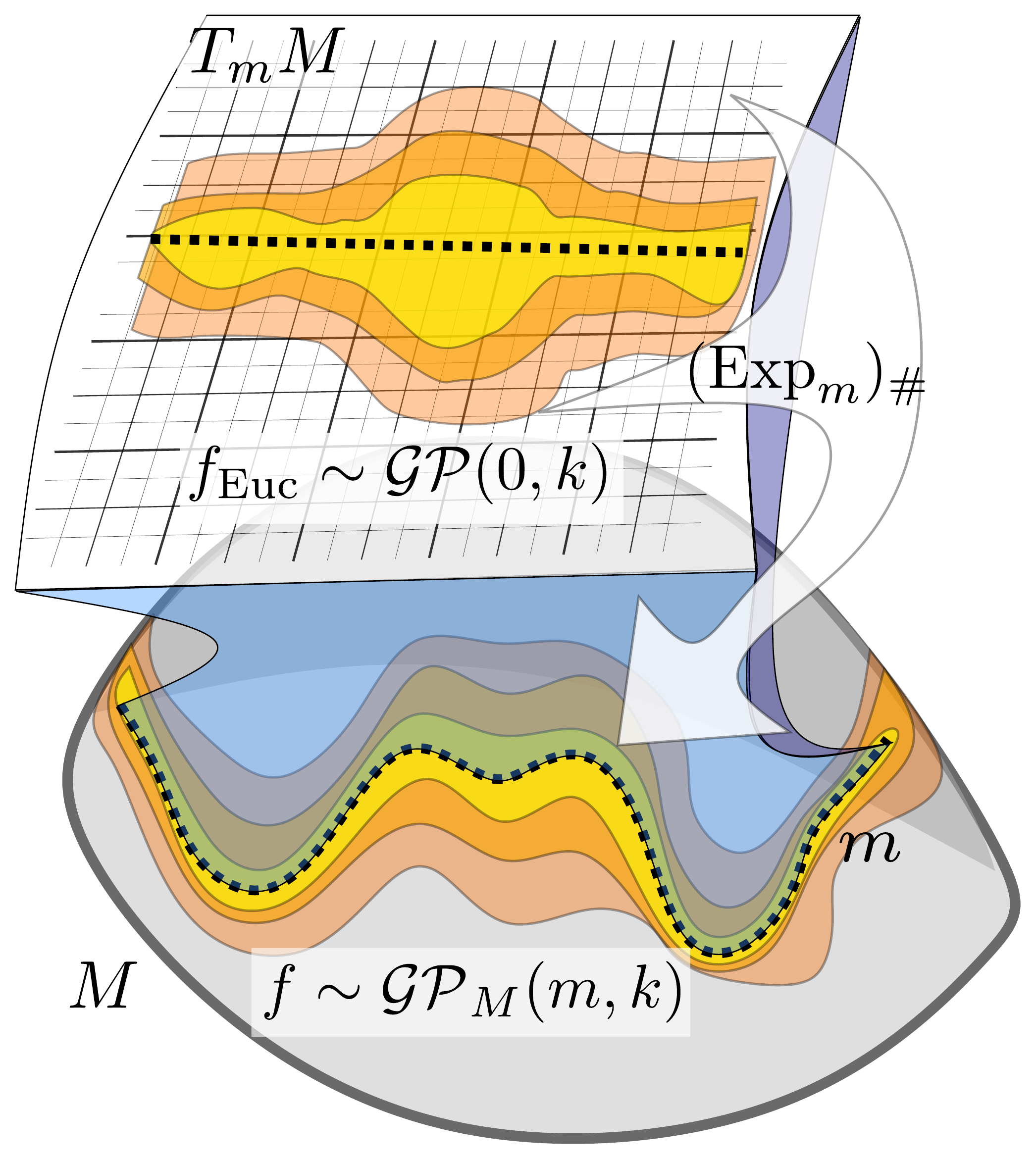}
\caption{A WGP $f$ can be viewed as defining a GP $f_{\mathrm{Euc}}$ in the tangent spaces $T_mM\subset M$ over the basepoint function, so that each marginal $f(x_i)$ is pushed-forward onto $M$ by $(\Exp_{m(x_i)})_\#(f(x_i))$.} 
\label{WGPIllustration}
\end{figure}
 Wrapped Gaussian processes generalize GPs to Riemannian manifolds~\citep{mallasto:cvpr:2018}. A collection $f$ of random points on a Riemannian manifold $M$ indexed over a set $\Omega$ is a \emph{wrapped Gaussian process} (WGP), if every finite subcollection $(f(\omega_i))_{i=1}^N$ is jointly WGD on $M^N$. The functions
\begin{equation}
\begin{aligned}
m(\omega) &= \mu_{\N_M}(f(\omega)),\\
k(\omega,\omega') &= \Cov_{\N_{M}}(f(\omega),f(\omega')),
\end{aligned}
\end{equation}
are called the \emph{basepoint function} and the \emph{tangent space covariance function} of $f$ (also called as kernel of $f$), respectively. To denote such a WGP, we use the notation $f\sim \GP_M(m,k)$.

Formally, a WGP $f$ can be viewed as a GP $f_{\mathrm{Euc}}$ on $T_mM\subset TM$, the family of tangent spaces over the basepoint function $m$. Then, the resulting GP is pushed forward to $M$ using the Riemannian exponential map $\Exp_m$ over $m$ to obtain the WGP, see Fig.~\ref{WGPIllustration}.

\begin{figure}[!htb]
\center
\includegraphics[width = 0.35\textwidth]{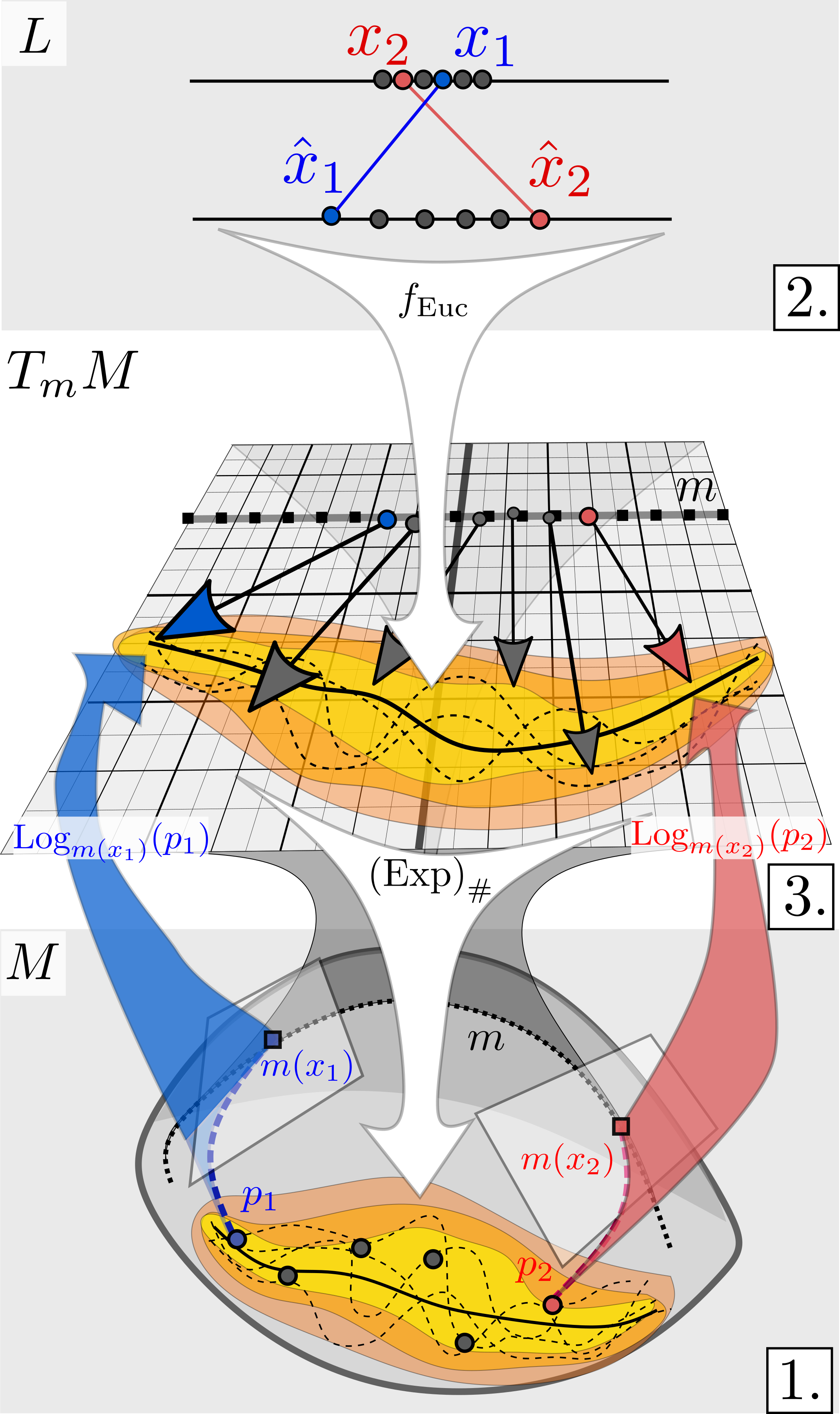}
\caption{\textbf{The WGPLVM pipeline.} \textbf{1.} The data $p_i\in \mathcal{M}$ (blue and red dots) is transformed to the tangent bundle by $p_i\mapsto \Log_{m(x_i)}(p_i)\in T_{m(x_i)}\mathcal{M} \subset T_m\mathcal{M}$ along the prior basepoint function $m$ (dotted black line) at initial latent variables $x_i$. \textbf{2.} A GPLVM is learned, yielding the latent variables $\hat{x}_i \in L$ and the GP $f_{\mathrm{Euc}}$ from $L$ to the tangent bundle. \textbf{3.} The GP $f_{\mathrm{Euc}}$ is then pushed forward onto $\mathcal{M}$ by $(\Exp)_\#(f_{\mathrm{Euc}})$, resulting in the predicted data submanifold.}
\vspace{-0.2in}
\label{WGPLVM}
\end{figure}

\section{WRAPPED GAUSSIAN PROCESS LATENT VARIABLE MODEL}
We now introduce the \emph{wrapped Gaussian process latent variable model} (WGPLVM) for data $P = \{p_i\}_{i=1}^N$ lying on an $n$-dimensional ambient Riemannian manifold $\mathcal{M}$. The goal of WGPLVM is to learn a lower-dimensional submanifold $M_{Pred} \subset \mathcal{M}$, where the data is assumed to reside. The WGPLVM model is a straight-forward generalization of the GPLVM model, where instead of GPs, we maximize the likelihood of our data combined with the latent variables under the WGPs that are suitable for the manifold context. The WGPLVM pipeline is illustrated in Fig.~\ref{WGPLVM}.

We consider a family of WGPs $f\sim \GP_\mathcal{M}(m ,k_\theta)$ from the latent space $L$ onto the ambient manifold $\mathcal{M}$, where $\theta\in \Theta$ are hyperparameters, that will be optimized over. The basepoint function $m$ can be utilized to delocalize the learning process in order to avoid distortions of the metric caused by linearization of the curved $\mathcal{M}$. The kernel $k_\theta$ affects how observations in different tangent spaces affect each other. For coherence, the kernel should be adapted to a smooth \emph{frame} (a smoothly changing basis over $m$). Such a frame can e.g.~be constructed by \emph{parallel transporting} a basis along $m$.

\textbf{The likelihood} assigned by the prior $f$ to a data point $p$ with associated latent variable $x$ is
\begin{equation}
\begin{aligned}
\prob\{p| x, \theta\} &= \sum_{v\in \Exp_{m(x)}^{-1}(p)}\N(v|\vect{0},K_{x,\theta})\\
&\approx \N\left(\Log_{m(x)}(p)|\vect{0},K_{x,\theta}\right) ,
\end{aligned}
\label{approxLikelihood}
\end{equation}
where $(K_{x,\theta})_{ij} = k_\theta(x^i,x^j)$ and $x= (x^1,x^2,...,x^n)$.

\textbf{The approximation} in Eq. \eqref{approxLikelihood} only takes into account the preimage of $p$ with a minimal norm (and thus maximal likelihood), denoted by $\Log_{m(x)}(p)$. The expression gives a lower bound for $\prob\{p| x, \theta\}$, thus, maximizing the likelihood of $\Log_{m(x)}(p)$  maximizes the lower bound for $\prob\{p| x, \theta\} $. It also enforces the WGPLVM to prefer \emph{local} models over ones that wrap considerably around the manifold. Note that, for manifolds with empty cut-locus (such as ones with non-positive curvature), the approximation in \eqref{approxLikelihood} is exact.

\textbf{The objective function} to be maximized is then the approximated log-likelihood 
\begin{equation}
\begin{aligned}
\ln\left(\prob\{p| x, \theta\}\right) \approx&  -\frac{dN}{2}\ln(2\pi)
  - \frac{d}{2}\ln |K_{x,\theta}|\\
  -& \frac{1}{2}\Log_{m(x)}(p)^TK_{x,\theta}^{-1}\Log_{m(x)}(p),
  \end{aligned}
\label{marginalLH}
\end{equation}
for which the gradient with respect to $x$ is given by
\begin{equation}
\begin{aligned}
&\frac{\partial}{\partial x_j} \ln\left(\prob\{p|x, \theta\}\right) \approx\\
 -&\frac{d}{2}\Tr\left(K_{x,\theta}^{-1}\frac{\partial K_{x,\theta}}{\partial x_j}\right)\\
-&\frac{1}{2}\Log_{m(x)}(p)^TK_{x,\theta}^{-1}D_{m(x)}\Log_{m(x)}(p)\frac{\partial m}{\partial x_j}(x)\\
-&\frac{1}{2}\Log_{m(x)}(p)^T\frac{\partial K_{x,\theta}^{-1}}{\partial x_j}\Log_{m(x)}(p),
\end{aligned}
\end{equation}
The differential $D_{m(x)}\Log_{m(x)}(p)$ can be computed using Jacobi fields as explained in expression \eqref{logcovder}, if no analytical expression exists.

Assuming that the data is i.i.d, the approximate log-likelihood for the data set $P$ can be written using Eq.~$\eqref{marginalLH}$, by considering $P$ as a single element of the product manifold $P^N$. This quantitity is then maximized by optimizing over the latent variables and the hyperparameters $\theta$, resulting in the optimal latent variables $\hat{X}$ and hyperparameters $\hat{\theta}$ for the kernel.

\textbf{The approximate submanifold} can then be predicted at arbitrary latent variables $X_{\mathrm{Pred}}$, by conditioning $\hat{f}\sim\GP_\mathcal{M}(m, k_{\hat{\theta}})$ on the data $P$ with the associated latent variables $\hat{X}$ (using Eq.~\eqref{cd_wgd}). The conditional distribution will then be a non-centered GP $f_{\mathrm{Euc}}\sim \GP(m_{\mathrm{Euc}}, k_{\mathrm{Euc}})$ defined on $T_m\mathcal{M}$ pushed forward by the exponential map (see Fig.~\ref{WGPLVM}), resulting in the predictive distribution $\varphi_{\pred} \sim (\Exp_{m(x)})_\#(f_{\mathrm{Euc}})$. Then, the \emph{mean prediction} is given by $\bar{\varphi}_{\pred}(x) = (\Exp_{m(x)})_\#(m_{\mathrm{Euc}})(x))$.  

In Eq. \eqref{cd_wgd}, if the preimage $\Exp_{\mu_2}^{-1}(p_2)$ is not uniquely defined, the conditional distribution is approximated by using a preimage with minimal norm, as previously. This approximation is justified as the weights $\lambda_v$ of the components of the Gaussian mixture decrease exponentially as $\|v\|_{p_2}$ increases.

\textbf{The initial latent variables} $X = \{x_i\}_{i=1}^N$ can be chosen strategically to aid optimization. We use \emph{principal geodesic analysis} (PGA)~\citep{fletcher04} and \emph{principal curves} \citep{principalcurves}. PGA is appropriate when the data expresses a geodesic trend (analogy of linearity on Riemannian manifolds), which is not the case for the femur dataset, see Fig.~\ref{SphereWGPLVM} in Section~\ref{sec:exp}.

\textbf{The Computational complexity} for the method is $\mathcal{O}(NL+ N^3)$, where $L$ is the cost of computing the Riemannian logarithm. This varies from manifold to manifold, but for example, in Section \ref{sec:exp}, the most expensive is $\mathcal{O}(d^3)$ for the Log-Euclidean metric on $d \times d$ symmetric, positive-definite matrices.

We provide a pseudo-algorithm for the method in the supplementary material.

\section{EXPERIMENTS} \label{sec:exp}
\begin{figure*}[ht]
\centering
\includegraphics[scale=0.45]{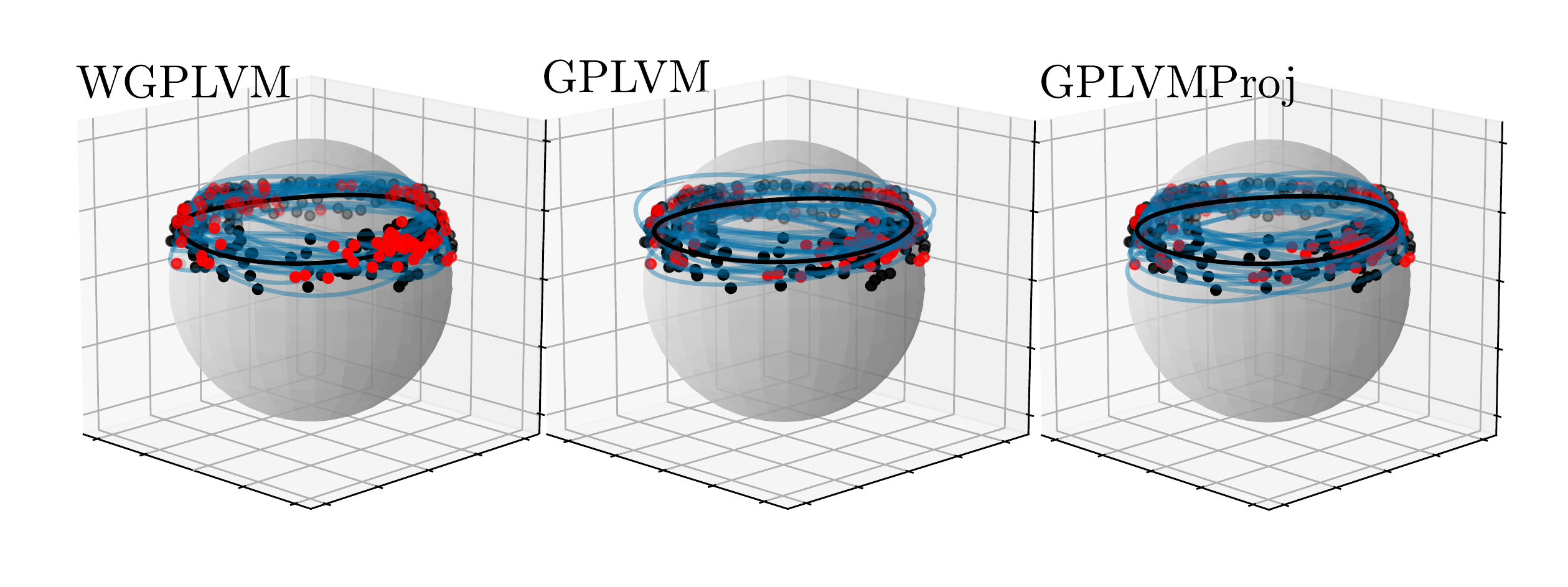}
\vspace{-0.35in}
\caption{WGPLVM, GPLVM and GPLVMProj submanifold predictions for the \emph{femur} data set. Mean predictions are in black, with 20 samples from the noise models (in blue). Training data in black, with test points in red.}
\label{SphereWGPLVM}
\end{figure*}
The WGPLVM is demonstrated on three different manifolds, arising from three different applications: The sphere, Kendall's shape space \citep{kendall}, and the space of symmetric, positive definite (SPD) matrices. Furthermore, the WGPLVM is compared with the Euclidean GPLVM, whose predictive distribution is expected not to lie on the manifold. This effect is clearly visible in Fig.~\ref{SphereWGPLVM}. A third model, also shown in Fig.~\ref{SphereWGPLVM}, is a modification of the Euclidean GPLVM, where the GP predictions are projected onto the manifold in order to make them satisfy the desired constraints.

We first introduce the datasets and their associated tasks, along with dataset-specific details related to training the models. In each case, we train the model assuming independent coordinates, applying the same kernel to each coordinate.

{\bf Femur dataset on $S^2$.} A set of directions $P=\{p_i\}_{i=1}^N \in S^2$ of the left \emph{femur} bone of a person walking in a circular pattern~\citep{mocapdata, principalcurves} is measured at $N=338$ time points. The movement is expected to be one dimensional and periodic, and thus we learn a $1$-dimensional submanifold homeomorphic to a circle to approximate the data manifold. The latent variable optimization is initialized using principal curves~\citep{principalcurves}, and the prior WGP and GP had kernel
\begin{equation}
k(t,t') = \sigma^2 \exp\left(-\frac{2\sin^2(|t-t'|)/2}{l^2}\right),
\end{equation}
and mean $m(t) = \mu_{S^2}$ and $m(t) = 0$, respectively, where $\mu_{S^2}$ is the \emph{Fr\'echet mean} of the training set and $\sigma^2, l^2$ are hyperparameters optimized to maximize the likelihood of the dataset $P$ with the latent variables $X$. The trained models are visualized in Fig.~\ref{SphereWGPLVM}.

{\bf Diatom shapes in Kendall's shape space.} Diatoms are unicellular algae, whose species are related to their shapes. In Kendall's shape space $M_K$ we analyze a set of outline shapes of 780 \emph{diatoms}~\citep{diatom, diatoms2} from 37 different species. For visualization, a two dimensional latent space is learned, using the prior $f\sim \GP_{M_K}(m, k)$, with constant basepoint function $m(t) = \mu_{M_K}$ set to be the Fr\'echet mean of the population and $k$ given by the \emph{radial basis function} (RBF) kernel
\begin{equation}
k(x,x') = \sigma^2 \exp\left(-\frac{\|x-x'\|_2^2}{2l^2}\right).
\end{equation}
We initialize the GPLVM and WGPLVM models with PGA and PCA, respectively.

{\bf Diffusion tensors in $SPD(3)$.} In the space of $3 \times 3$ SPD matrices with the Log-Euclidean metric~\citep{arsigny06}, we collect a set of $750$ diffusion tensors from a diffusion MRI dataset, sampled with approximately uniform fractional anisotropy (FA) values. The FA is a well-known tensor shape descriptor; see the supplementary material for the definition. As SPD matrices form an open subset of the Euclidean space of symmetric matrices, \emph{we do not get a \enquote{for free} dimensionality reduction} by restricting to SPD matrices. Instead, the data is transformed nonlinearly according to the Log-Euclidean metric,  which is commonly used for diffusion tensors~\citep{arsigny06}. The diffusion MRI image was a single subject from the Human Connectome Project~\citep{hcp1,hcp2,Sotiropoulos2013}. In diffusion MRI, low-dimensional encoding with uncertainty estimates
may speed up image acquisition and processing.

\begin{table*}[h!]
\centering
 \begin{tabular}{c||c|c|c|c}
 \hline
  Riemannian & Femur & Diatoms & Diffusion tensors & Crypto-tensors\\
  \hline
  \hline
   GPLVMProj &$(9.22 \pm 0.55) \times 10^{-2}$  & $(2.48 \pm 0.25)\times 10^{-2}$ & $0.582 \pm 0.025$ & $21.91 \pm 2.26$\\
   WGPLVM & $\mathbf{(9.20 \pm 0.53) \times 10^{-2}}$ &$\mathbf{(2.39 \pm 0.15)\times 10^{-2}}$& $\mathbf{0.391 \pm 0.035}$ & $\mathbf{3.04 \pm 0.26}$\\
  \hline
 \end{tabular}
  \begin{tabular}{c||c|c|c|c}
 \hline
  Euclidean & Femur & Diatoms & Diffusion tensors & Crypto-tensors\\
  \hline
  \hline
   GPLVM & $(9.21 \pm 0.55) \times 10^{-2}$ &$(2.48 \pm 0.25)\times 10^{-2}$ & $\mathbf{(6.03 \pm 0.34)\times 10^{-2}}$ & $(7.36 \pm 5.27)\times 10^5$\\
   GPLVMProj &$(9.21 \pm 0.55) \times 10^{-2}$  & $(2.48 \pm 0.25\times 10^{-2}$ & $\mathbf{(6.03 \pm 0.34)\times 10^{-2}}$ & $(5.49 \pm 3.17)\times 10^5$\\
   WGPLVM & $\mathbf{(9.19 \pm 0.53) \times 10^{-2}}$ &$\mathbf{(2.39 \pm 0.15)\times 10^{-2}}$& $(7.54 \pm 0.36)\times 10^{-2}$ & $(8.69 \pm 7.12)\times 10^5$\\
  \hline
 \end{tabular}
\caption{Mean $\pm$ standard error of mean reconstruction errors, measured in RMSE, over $10$ repetitions of the experiment. \textbf{Top table:} Deviations measured in the intrinsic distance on the manifold. \textbf{Bottom table:} Deviations measured in the Euclidean distance.}
\label{table_recon}
\end{table*}

{\bf Crypto-tensors in $SPD(10)$.} On $SPD(10)$ we collect the price of 10 popular crypto-currencies\footnote{Bitcoin, Dash, Digibyte, Dogecoin, Litecoin, Vertcoin, Stellar, Monero, Ripple, and Verge.} in the time 2.12.2014-15.5.2018. The crypto-currency intra-relationship at a given time is encoded in the covariance matrix between the prices in the past 20 days; we include every $7$th day in the period, resulting in $126$ $10 \times 10$ covariance matrices. \citet{GeneralizedWishartProcesses} provide a discussion of covariance descriptors in economy. As the acquired covariance matrices in $SPD(10)$ have eigenvalues in different orders of magnitude, we use the Log-Euclidean metric~\citep{arsigny06}, capturing this trend better.

For both $SPD(n)$ datasets, the basepoint function, the kernel and the latent variable initialization are chosen as for Kendall's shape space. The latent spaces are chosen to be $2$-dimensional for visualization purposes.

\textbf{Application 1: Encoding.} The datasets are divided  into training and test sets (consisting of $\sfrac{8}{10}$ and $\sfrac{2}{10}$ of the data, respectively), and we learn the models $\varphi_{\pred}$ on the training set. Each test element $p$ is \enquote{encoded} by the projection $\pi\colon p\mapsto \operatorname*{argmax}_{x \in L} \mathbb{P}\{\varphi_{\mathrm{pred}}(x) = p\}$. We assess the quality of this encoding by measuring the root-mean-square error (RMSE) of the reconstruction,  where the error is measured both by the Euclidean metric and the intrinsic metric. Each experiment was repeated 10 times with different training and test sets; the results are reported in Table~\ref{table_recon}.

Under the intrinsic metric, the WGPLVM performs significantly better on the tensor datasets, and marginally better in the two other cases. Under the Euclidean metric the WGPLVM encoding is better in two cases, worse in one, and inconclusive for the crypto-tensors where no model is significantly better than the others.
\begin{figure}[t]
\includegraphics[width=0.45\linewidth]{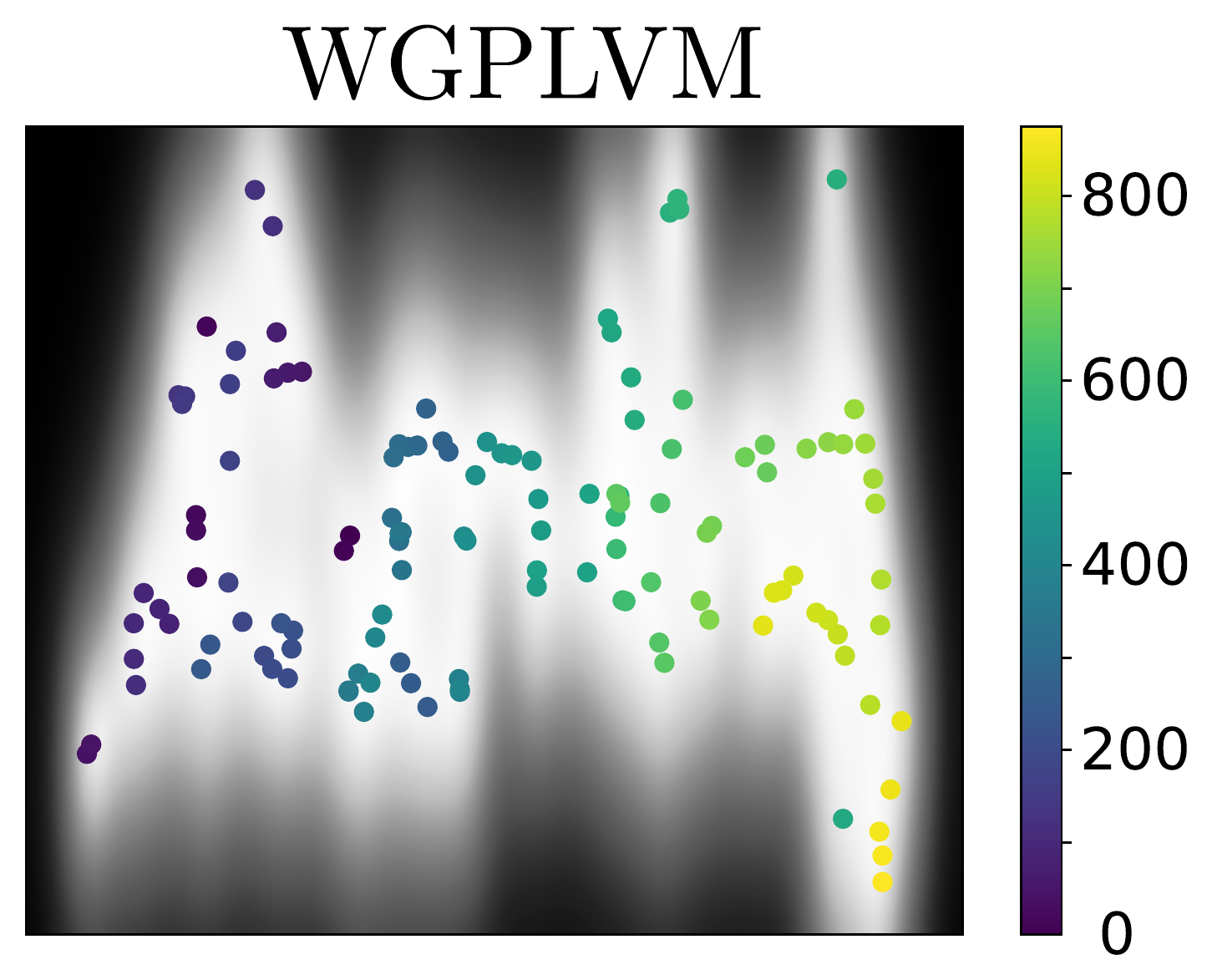} \hfil \includegraphics[width=0.45\linewidth]{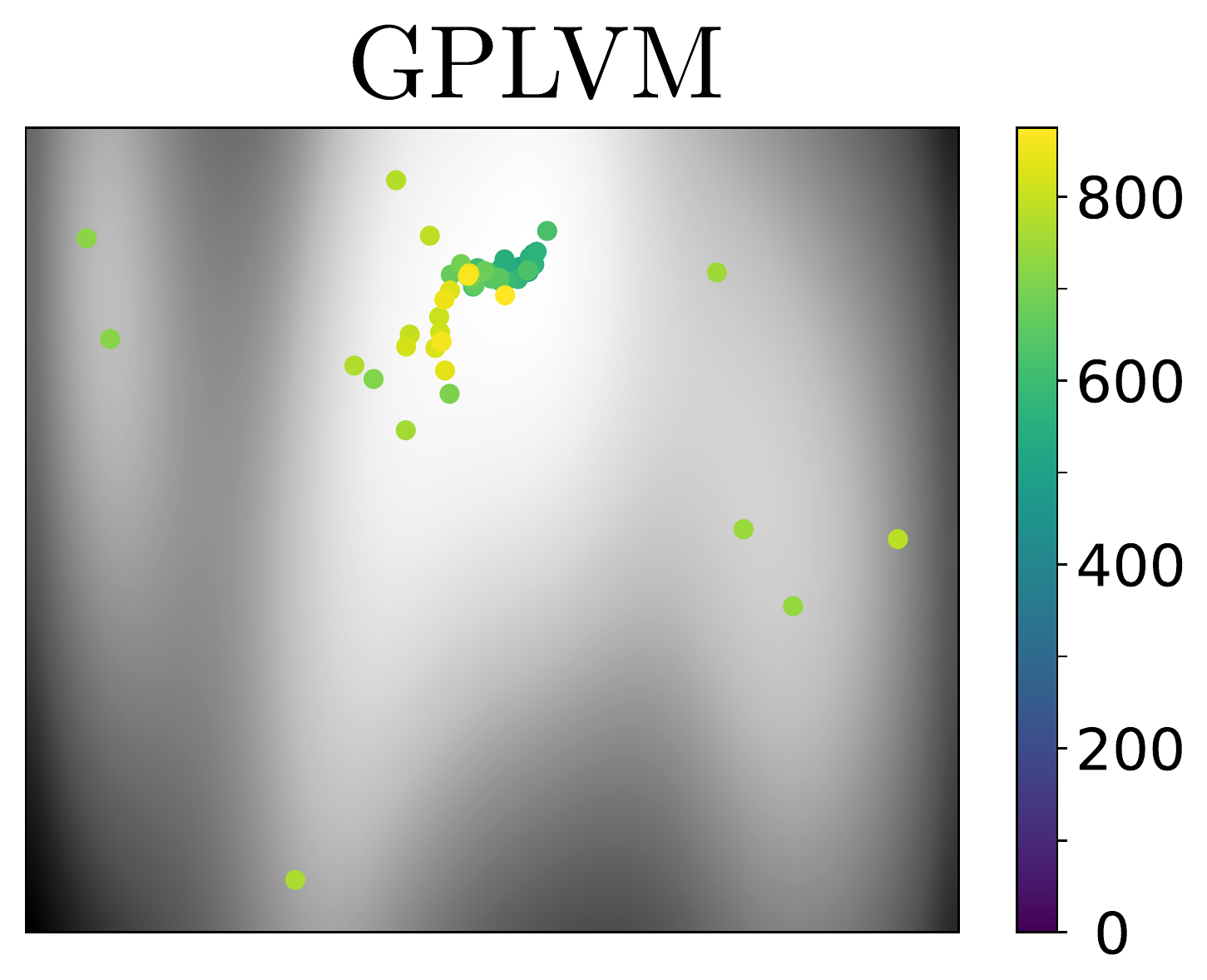}
\caption{The latent space for the crypto-tensor dataset, with days visualized by color. Note that for GPLVM, the dark blue points corresponding to early times are hidden underneath the green points.}
\label{fig:latent_bitcoin}
\vspace{-0.1in}
\end{figure}

\begin{figure*}
\centering
\includegraphics[width = \textwidth]{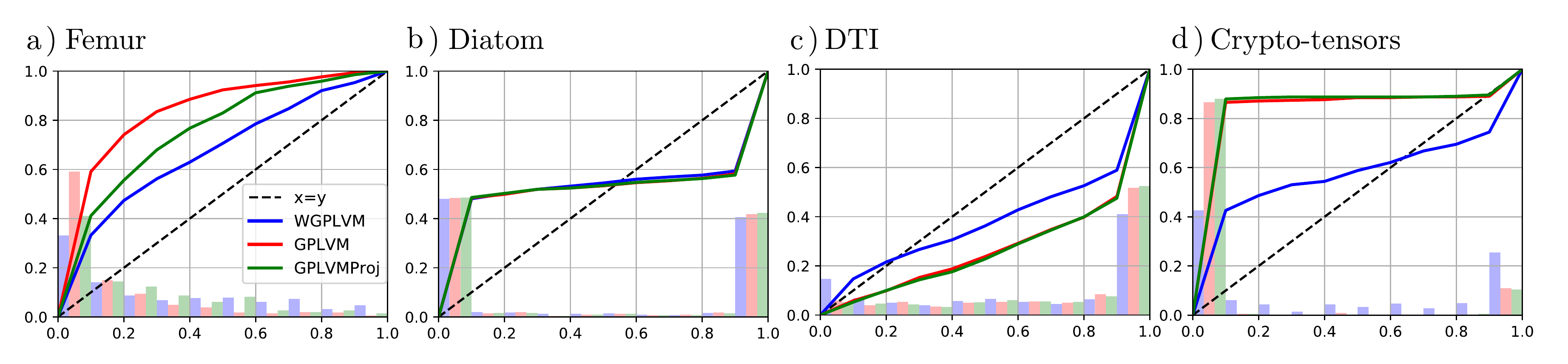}
\vspace{-0.3in}
\caption{Uncertainty estimates given by the WGPLVM, GPLVM and projected GPLVM models for the four datasets. The bars represent the frequency of occurances, where the fraction of samples, given by the x-value, lie closer to the mean prediction than a test point. The continuous curves represent the cumulative distributions. Whenever the cumulative distribution lies above $x=y$, we are overestimating the corresponding quantile.}
\vspace{-0.1in}
\label{sphereConfidence}
\end{figure*}
\textbf{Application 2: Uncertainty quantification.} Importantly, GPLVM learns a probabilistic model, producing an estimate of uncertainty. We evaluate these uncertainty estimates on all four datasets. Since the predictive distributions live in different spaces, the likelihoods of observed data under the different models are not directly comparable. However, all three models yield confidence intervals, which we compare using 10 resampled training and test sets (\sfrac{8}{10} and \sfrac{8}{10} of the data). The test set is projected onto the predicted submanifold via $\pi$. Then, we sample the respective predictive distributions $50$ times, computing the fraction of samples closer to the mean prediction than the test point. The results are visualized in Fig.~\ref{sphereConfidence}, where the densities of these fractions are shown with corresponding cumulative distributions. For a perfect model fit, we would observe the $x=y$ curve (dashed line) as the cumulative distribution. The experiment shows that all models estimate uncertainty incorrectly, but that WGPLVM obtains the best estimate.

\textbf{Application 3: Visualization.} In Fig.~\ref{fig:latent_bitcoin}, we illustrate the latent spaces of WGPLVM versus GPLVM on the crypto-tensor dataset, which comes with an associated time variable, shown in color. The WGPLVM provides a smoother and more consistent transition in color, while the GPLVM plots all the earlier (dark blue) tensors on top of each other. Similar visualizations for the other datasets can be found in the supplementary material; in these examples, the two visualizations are not significantly different in quality.

In the supplementary material, we provide a discussion on why our model might perform better in the $SPD(n)$ experiments, including a comparison between the Euclidean and Riemannian geometries.

\section{DISCUSSION AND CONCLUSION}
We introduced the WGPLVM for non-parametric and probabilistic submanifold learning on Riemannian manifolds. The model encodes known constraints or invariances, and provides model flexibility, as metrics other than the Euclidean one can be incorporated. This is useful if a different metric captures trends in the data better. The model was evaluated on several manifolds and tasks against the GPLVM and a modified GPLVM, which projects predictions onto the manifold. 

The experimental results show that the WGPLVM provides a better probabilistic model to fit the data; in particular the uncertainty estimates are superior to the Euclidean models on three out of four datasets, and virtually identical on the fourth. We note that for Euclidean models, the uncertainty is visibly higher. These are strong indications that our model carries out modelling the data distribution better. The mean predictions of the WGPLVM encode the data space significantly better than the GPLVM and projected GPLVM models on two of the datasets, and marginally better on the other two, when measured in the Riemannian metric. Under the Euclidean metricr, the GPVLM performs notably better in one experiment, and WGPLVM marginally better in two. On crypto-tensors, we deem the results inconclusive due to high variance. The aforementioned effects are also seen in the latent space visualizations, e.g.~on the cryptotensors the WGPLVM better detects small-scale differences in the early time steps.

One might suspect that the improved performance stems from a \enquote{for free} dimensionality reduction through constraints. However, we note that the most significant improvement in both reconstruction error and visualization was obtained on $SPD(n)$, where 
the Riemannian manifold is a full-dimensional, convex subset of the Euclidean ambient space. This might still be due to the constraints, which forces the distributions to lie in the manifold. The difference could also be caused by the choice of metric. For the crypto-tensors in particular, we observe that some of the eigenvalues are very small; the Log-Euclidean metric essentially acts as a log-transform and therefore converts the data to a scale on which changes in the smaller eigenvalues can be detected.

In three of the experiments, the mean predictions of GPLVM lie essentially on the manifold, thus the projected version does not improve the mean reconstruction error. However, in the femur experiment, the uncertainty estimates are clearly improved, but also notably outperformed by WGPLVM. Due to the metric and curvature of the manifold, interpolation between two points in the ambient space $\R^n$ does not necessarily project even closely onto the manifold interpolation between the projected points. This distortion affects the statistics relying on interpolation, and explains both the reduced reconstruction capability and the increased variance. Furthermore, the projected model ignores any metric choices imposed on the manifold.

Although the WGPLVM provides flexibility through the prior basepoint function, we fixed this to be the Fr\'echet mean of the training set in our experiments. The choice is well justified if the data is local enough, and makes the comparison to GPLVM fair. The flexibility to 
delocalize the learning process through the basepoint function is, however, important for inference on manifolds when the locality assumption fails. The non-trivial optimization of the basepoint function thus provides a venue for future research.

In summary, the WGPLVM is a probabilistic submanifold learning algorithm that respects known Riemannian manifold structure in the data by taking values in the associated Riemannian manifold. We compare the model to its Euclidean counterparts on a number of manifolds, datasets and tasks, and show that it has superior representation capabilities more faithful visualizations and improved uncertainty estimates.

\subsubsection*{Acknowledgements}
AM and AF were supported by Centre for Stochastic Geometry and Advanced Bioimaging, funded by a grant from the Villum Foundation, and SH was supported by the European Research Council (ERC) under the European Union’s Horizon 2020 research and innovation programme (grant agreement n$^\circ$ 757360), as well as a research grant (15334) from VILLUM FONDEN.  Data were provided [in part] by the Human Connectome Project, WU-Minn Consortium (Principal Investigators: David Van Essen and Kamil Ugurbil; 1U54MH091657) funded by the 16 NIH Institutes and Centers that support the NIH Blueprint for Neuroscience Research; and by the McDonnell Center for Systems Neuroscience at Washington University. The data [in part] used in this project was obtained from \url{mocap.cs.cmu.edu}. The database was created with funding from NSF EIA-0196217. The authors wish to thank Thomas Hamelryck for helpful comments.

\bibliographystyle{plainnat}
\bibliography{egbib}

\begin{thebibliography}{38}
\providecommand{\natexlab}[1]{#1}
\providecommand{\url}[1]{\texttt{#1}}
\expandafter\ifx\csname urlstyle\endcsname\relax
  \providecommand{\doi}[1]{doi: #1}\else
  \providecommand{\doi}{doi: \begingroup \urlstyle{rm}\Url}\fi

\bibitem[Arsigny et~al.(2006)Arsigny, Fillard, Pennec, and Ayache]{arsigny06}
Vincent Arsigny, Pierre Fillard, Xavier Pennec, and Nicholas Ayache.
\newblock Log-euclidean metrics for fast and simple calculus on diffusion
  tensors.
\newblock \emph{Magnetic Resonance in Medicine: An Official Journal of the
  International Society for Magnetic Resonance in Medicine}, 56\penalty0
  (2):\penalty0 411--421, 2006.

\bibitem[Batchelor et~al.(2005)Batchelor, Moakher, Atkinson, Calamante, and
  Connelly]{batchelor}
Phillipp~G. Batchelor, Maher Moakher, David. Atkinson, Fernando. Calamante, and
  Alan Connelly.
\newblock A rigorous framework for diffusion tensor calculus.
\newblock \emph{Magnetic Resonance in Medicine}, 53\penalty0 (1):\penalty0
  221--225, 2005.
\newblock ISSN 1522-2594.

\bibitem[Belkin and Niyogi(2003)]{lapeigmaps}
Mikhail Belkin and Partha Niyogi.
\newblock Laplacian eigenmaps for dimensionality reduction and data
  representation.
\newblock \emph{Neural Computation}, 15\penalty0 (6):\penalty0 1373--1396,
  2003.

\bibitem[{CMU Graphics Lab}(2003)]{mocapdata}
{CMU Graphics Lab}.
\newblock {CMU Graphics Lab Motion Capture Database }.
\newblock \url{http://mocap.cs.cmu.edu/}, 2003.
\newblock The data used in this project was obtained from mocap.cs.cmu.edu. The
  database was created with funding from NSF EIA-0196217.

\bibitem[Do~Carmo(1992)]{docarmo}
Manfredo~Perdigao Do~Carmo.
\newblock \emph{{R}iemannian geometry}.
\newblock Birkhauser, 1992.

\bibitem[du~Buf and Bayer(2002)]{diatoms2}
Hans du~Buf and Micha Bayer.
\newblock \emph{Automatic Diatom Identification}.
\newblock 2002.

\bibitem[Fletcher and Joshi(2004)]{fletcherjoshi2004}
P.~Thomas Fletcher and Sarang Joshi.
\newblock \emph{Principal Geodesic Analysis on Symmetric Spaces: Statistics of
  Diffusion Tensors}, pages 87--98.
\newblock 2004.

\bibitem[Fletcher et~al.(2004)Fletcher, Lu, Pizer, and Joshi]{fletcher04}
P.~Thomas Fletcher, Conglin Lu, Stephen~M. Pizer, and Sarang Joshi.
\newblock Principal geodesic analysis for the study of nonlinear statistics of
  shape.
\newblock \emph{IEEE transactions on medical imaging}, 23\penalty0
  (8):\penalty0 995--1005, 2004.

\bibitem[Freifeld and Black(2012)]{Freifeld:ECCV:2012}
Oren Freifeld and Michael~J Black.
\newblock {L}ie bodies: A manifold representation of {3D} human shape.
\newblock In {A. Fitzgibbon et al. (Eds.)}, editor, \emph{European Conference
  on Computer Vision (ECCV)}, Part I, LNCS 7572, pages 1--14. Springer-Verlag,
  2012.

\bibitem[Glasser et~al.(2013)Glasser, Sotiropoulos, Wilson, Coalson, Fischl,
  Andersson, Xu, Jbabdi, Webster, Polimeni, et~al.]{hcp2}
Matthew~F. Glasser, Stamatios~N. Sotiropoulos, J.~Anthony Wilson, Timothy~S.
  Coalson, Bruce Fischl, Jesper~L. Andersson, Junqian Xu, Saad Jbabdi, Matthew
  Webster, Jonathan~R. Polimeni, et~al.
\newblock The minimal preprocessing pipelines for the {H}uman {C}onnectome
  project.
\newblock \emph{Neuroimage}, 80:\penalty0 105--124, 2013.

\bibitem[Goodfellow et~al.(2014)Goodfellow, Pouget-Abadie, Mirza, Xu,
  Warde-Farley, Ozair, Courville, and Bengio]{gans}
Ian Goodfellow, Jean Pouget-Abadie, Mehdi Mirza, Bing Xu, David Warde-Farley,
  Sherjil Ozair, Aaron Courville, and Yoshua Bengio.
\newblock Generative adversarial nets.
\newblock In \emph{Advances in Neural Information Processing Systems (NIPS)},
  2014.

\bibitem[Hauberg(2016)]{principalcurves}
S{\o}ren Hauberg.
\newblock Principal curves on {R}iemannian manifolds.
\newblock \emph{IEEE Transactions on Pattern Analysis and Machine Intelligence
  (TPAMI)}, 2016.

\bibitem[Hauberg et~al.(2012)Hauberg, Freifeld, and Black]{hauberg2012}
S{\o}ren Hauberg, Oren Freifeld, and Michael~J. Black.
\newblock A geometric take on metric learning.
\newblock In P.~Bartlett, F.C.N. Pereira, C.J.C. Burges, L.~Bottou, and K.Q.
  Weinberger, editors, \emph{Advances in Neural Information Processing Systems
  (NIPS) 25}, pages 2033--2041. MIT Press, 2012.

\bibitem[Huckemann et~al.(2010)Huckemann, Hotz, and Munk]{huckemann10}
Stephan Huckemann, Thomas Hotz, and Axel Munk.
\newblock Intrinsic shape analysis: Geodesic {PCA} for {R}iemannian manifolds
  modulo isometric {L}ie group actions.
\newblock \emph{Statistica Sinica}, pages 1--58, 2010.

\bibitem[Jalba et~al.(2006)Jalba, Wilkinson, and Roerdink]{diatom}
Andrei~C. Jalba, Michael~HF Wilkinson, and Jos~BTM Roerdink.
\newblock Shape representation and recognition through morphological curvature
  scale spaces.
\newblock \emph{{IEEE} Transactions on Image Processing}, 15\penalty0
  (2):\penalty0 331--341, feb 2006.

\bibitem[Kendall(1984)]{kendall}
David~G. Kendall.
\newblock Shape manifolds, {P}rocrustean metrics, and complex projective
  spaces.
\newblock \emph{Bulletin of the London Mathematical Society}, 16\penalty0
  (2):\penalty0 81--121, 1984.

\bibitem[Kingma and Welling(2014)]{autoencoders}
Diederik~P. Kingma and Max Welling.
\newblock Auto-{E}ncoding {V}ariational {B}ayes.
\newblock In \emph{Proceedings of the 2nd International Conference on Learning
  Representations (ICLR)}, 2014.

\bibitem[Lawrence(2004)]{lawrence04}
Neil~D Lawrence.
\newblock {G}aussian process latent variable models for visualisation of high
  dimensional data.
\newblock In \emph{Advances in neural information processing systems}, pages
  329--336, 2004.

\bibitem[Lawrence(2005)]{gplvm}
Neil~D. Lawrence.
\newblock Probabilistic non-linear principal component analysis with {G}aussian
  process latent variable models.
\newblock \emph{J. Mach. Learn. Res.}, 6:\penalty0 1783--1816, 2005.
\newblock ISSN 1532-4435.

\bibitem[Lawrence and Moore(2007)]{lawrence07}
Neil~D. Lawrence and Andrew~J. Moore.
\newblock Hierarchical {G}aussian process latent variable models.
\newblock In \emph{Proceedings of the 24th international conference on Machine
  learning}, pages 481--488. ACM, 2007.

\bibitem[Lin et~al.(2017)Lin, Niu, Cheung, and Dunson]{lin17}
Lizhen Lin, Mu~Niu, Pokman Cheung, and David Dunson.
\newblock Extrinsic gaussian processes for regression and classification on
  manifolds.
\newblock \emph{arXiv preprint arXiv:1706.08757}, 2017.

\bibitem[Mallasto and Feragen(2018)]{mallasto:cvpr:2018}
Anton Mallasto and Aasa Feragen.
\newblock Wrapped {G}aussian process regression on {R}iemannian manifolds.
\newblock In \emph{CVPR - IEEE Conference on Computer Vision and Pattern
  Recognition, to appear}, 2018.

\bibitem[Mardia and Jupp(2009)]{mardia2009directional}
Kanti~V. Mardia and Peter~E. Jupp.
\newblock \emph{Directional statistics}, volume 494.
\newblock John Wiley \& Sons, 2009.

\bibitem[Pennec(2015)]{barycentric}
Xavier Pennec.
\newblock Barycentric subspaces and affine spans in manifolds.
\newblock In \emph{Geometric Science of Information - Second International
  Conference, {GSI} 2015, Palaiseau, France, October 28-30, 2015, Proceedings},
  pages 12--21, 2015.

\bibitem[Pennec(2016)]{pennec16barycentric}
Xavier Pennec.
\newblock Barycentric subspace analysis on manifolds.
\newblock \emph{arXiv preprint arXiv:1607.02833}, 2016.

\bibitem[Rasmussen(2004)]{rasmussen}
Carl~Edward Rasmussen.
\newblock {G}aussian processes in machine learning.
\newblock In \emph{Advanced lectures on machine learning}, pages 63--71.
  Springer, 2004.

\bibitem[Roweis and Saul(2000)]{lle}
Sam~T. Roweis and Lawrence~K. Saul.
\newblock Nonlinear dimensionality reduction by locally linear embedding.
\newblock \emph{SCIENCE}, 290:\penalty0 2323--2326, 2000.

\bibitem[Sotiropoulos et~al.(2013)Sotiropoulos, Moeller, Jbabdi, Xu, Andersson,
  Auerbach, Yacoub, Feinberg, Setsompop, Wald, et~al.]{Sotiropoulos2013}
Stamatios~N. Sotiropoulos, Steen Moeller, Saad Jbabdi, Jungqian Xu, Jesper
  Andersson, Edward~John Auerbach, Essa Yacoub, David~A. Feinberg, Kawin
  Setsompop, Lawrence~L. Wald, et~al.
\newblock Effects of image reconstruction on fiber orientation mapping from
  multichannel diffusion {MRI}: reducing the noise floor using {SENSE}.
\newblock \emph{Magnetic resonance in medicine}, 70\penalty0 (6):\penalty0
  1682--1689, 2013.

\bibitem[Tenenbaum et~al.(2000)Tenenbaum, Silva, and Langford]{isomap}
Joshua~B. Tenenbaum, Vin~de Silva, and John~C. Langford.
\newblock A global geometric framework for nonlinear dimensionality reduction.
\newblock \emph{Science}, 290\penalty0 (5500):\penalty0 2319--2323, 2000.

\bibitem[Titsias and Lawrence(2010)]{titsias10}
Michalis Titsias and Neil~D. Lawrence.
\newblock {B}ayesian {G}aussian process latent variable model.
\newblock In \emph{Proceedings of the Thirteenth International Conference on
  Artificial Intelligence and Statistics}, pages 844--851, 2010.

\bibitem[Tuzel et~al.(2006)Tuzel, Porikli, and Meer]{Tuzel:ECCV:2006}
Oncel. Tuzel, Fatih Porikli, and Peter Meer.
\newblock Region covariance: A fast descriptor for detection and
  classification.
\newblock \emph{European Conference on Computer Vision (ECCV)}, pages 589--600,
  2006.

\bibitem[Urtasun and Darrell(2007)]{urtasun07}
Raquel Urtasun and Trevor Darrell.
\newblock Discriminative {G}aussian process latent variable model for
  classification.
\newblock In \emph{Proceedings of the 24th international conference on Machine
  learning}, pages 927--934. ACM, 2007.

\bibitem[Urtasun et~al.(2006)Urtasun, Fleet, and Fua]{urtasun20063d}
Raquel Urtasun, David~J. Fleet, and Pascal Fua.
\newblock 3d people tracking with {G}aussian process dynamical models.
\newblock In \emph{Computer Vision and Pattern Recognition, 2006 IEEE Computer
  Society Conference on}, volume~1, pages 238--245. IEEE, 2006.

\bibitem[Urtasun et~al.(2008)Urtasun, Fleet, Geiger, Popovi{\'c}, Darrell, and
  Lawrence]{urtasun08}
Raquel Urtasun, David~J. Fleet, Andreas Geiger, Jovan Popovi{\'c}, Trevor~J.
  Darrell, and Neil~D Lawrence.
\newblock Topologically-constrained latent variable models.
\newblock In \emph{Proceedings of the 25th international conference on Machine
  learning}, pages 1080--1087. ACM, 2008.

\bibitem[Van~Essen et~al.(2013)Van~Essen, Smith, Barch, Behrens, Yacoub,
  Ugurbil, Consortium, et~al.]{hcp1}
David~C. Van~Essen, Stephen~M. Smith, Deanna~M. Barch, Timothy~EJ Behrens, Essa
  Yacoub, Kamil Ugurbil, Wu-Minn~HCP Consortium, et~al.
\newblock The wu-minn {H}uman {C}onnectome project: an overview.
\newblock \emph{Neuroimage}, 80:\penalty0 62--79, 2013.

\bibitem[Varol et~al.(2012)Varol, Salzmann, Fua, and Urtasun]{varol12}
Aydin Varol, Mathieu Salzmann, Pascal Fua, and Raquel Urtasun.
\newblock A constrained latent variable model.
\newblock In \emph{Computer Vision and Pattern Recognition (CVPR), 2012 IEEE
  Conference on}, pages 2248--2255. Ieee, 2012.

\bibitem[Wilson and Ghahramani(2011)]{GeneralizedWishartProcesses}
Andrew~Gordon Wilson and Zoubin Ghahramani.
\newblock Generalised {W}ishart processes.
\newblock In \emph{{UAI} 2011, Proceedings of the Twenty-Seventh Conference on
  Uncertainty in Artificial Intelligence, Barcelona, Spain, July 14-17, 2011},
  pages 736--744, 2011.

\bibitem[Zhang and Fletcher(2013)]{ppga}
Miaomiao Zhang and P.~Thomas Fletcher.
\newblock Probabilistic principal geodesic analysis.
\newblock In \emph{Advances in Neural Information Processing Systems 26: 27th
  Annual Conference on Neural Information Processing Systems 2013. Proceedings
  of a meeting held December 5-8, 2013, Lake Tahoe, Nevada, United States.},
  pages 1178--1186, 2013.

\end{thebibliography}

\renewcommand{\thesubsection}{\Alph{subsection}}
\appendix
\section*{Supplementary Material}
\subsection{Pseudo-Algorithm forWGPLVM}
A pseudo-code algorithm for training the WGPLVM is provied in Alg. \ref{alg:WGPLVM}.
\begin{algorithm}[tb]
   \caption{Training WGPLVM. Input: basepoint function $m$, kernel $k_\Theta$, initial latent variables $x=\{x_i\}_{i=1}^N$, dataset $p=\{p_i\}_{i=1}^N$, learning rate $\lambda$. Each logarithmic map should be express with respect to a frame $W$ on the manifold.}
\begin{algorithmic}
	\While{Not converged}
	\State \emph{$\#$ Compute logarithmic maps and save into a matrix as rows}
	\State$[\Log_{m(x)}(p)]_i\leftarrow \Log_{m(x_i)}(p_i)$
	\State \emph{$\#$ Compute prior covariance matrix:}
	\State $[K_{x,\Theta}]_{ij} \leftarrow k_\Theta(x_i,x_j)$
	\State \emph{$\#$ Compute objective:}
	\State $L \leftarrow --\frac{dN}{2}\ln(2\pi) - \frac{d}{2}\ln |K_{x,\Theta}|- \frac{1}{2}\Log_{m(x)}(p)^TK_{x,\Theta}^{-1}\Log_{m(x)}(p)$
	\State \emph{$\#$ Compute gradients and update parameters}
	\State $x \leftarrow x + \lambda \nabla_x L$
	\State $\Theta \leftarrow \Theta + \lambda \nabla_\Theta L$
	\EndWhile
\end{algorithmic}
\label{alg:WGPLVM}
\end{algorithm}

\subsection{Details on Manifolds Used}
{\bf The $n$-sphere} $S^n$ is a Riemannian manifold with exponential and logarithmic maps given by
\begin{align}
\begin{aligned}
\Exp_p(v) &= \cos(\|v\|_2)p + \sin(\|v\|_2)\frac{v}{\|v\|_2},\\
\Log_p(q) &=  \arccos{(\langle p, q\rangle)}\frac{q - \langle p,q\rangle p}{\|q-\langle p, q\rangle p\|_2},
\end{aligned}
\end{align}
where $\|\cdot\|_2$ is the $2$-norm induced by the standard Euclidean innerproduct $\langle\cdot,\cdot\rangle$.

{\bf Kendall's shape space} forms a quontient manifold of the sphere, so the operations defined for $S^n$ apply, when working with the right quotient representatives. Kendall's shape space has the additional constraint of representing shapes with respect to an optimal translation between a pair of shapes. Let $X,Y$ be the $2\times N$ data matrices of two shapes, where $N$ is the amount of landmarks, and each column represents the $x,y$-coordinates after quontienting away scale and translation. Then, the \emph{Procrustean} distance between the shapes $X,Y$ is given by
\begin{equation}
\min\limits_R\|X-RY\|_2,
\end{equation}
where $R$ is a rotation matrix. The shapes are aligned by choosing a reference point, and aligning the population elements by minimizing the Procrustean distance.

{\bf The space $SPD(n)$ of symmetric, positive definite matrices} can be given the structure of a Riemannian manifold, by endowing it with the \emph{Log-Euclidean} metric. The tangent space at each point is the space of $n$-by-$n$ symmetric matrices, and the affine-invariant metric is given by
\begin{equation}
g_P(V,U) = \mathrm{Trace}[V^TU],
\end{equation}
and the exponential and logarithmic maps are given by
\begin{equation}
\Exp_P(A) = \exp(\log(p) + v),~~\Log_P(Q) = \log(Q)-\log(P),
\end{equation}
where $\exp$ stands for the matrix exponential and $\log$ for the matrix logarithm.

\newpage 

\subsection{Latent Space Visualization}

Here we provide the latent space visualizations for the diffusion-tensor and diatom datasets.

\begin{figure}[h!]
\includegraphics[width=0.45\linewidth]{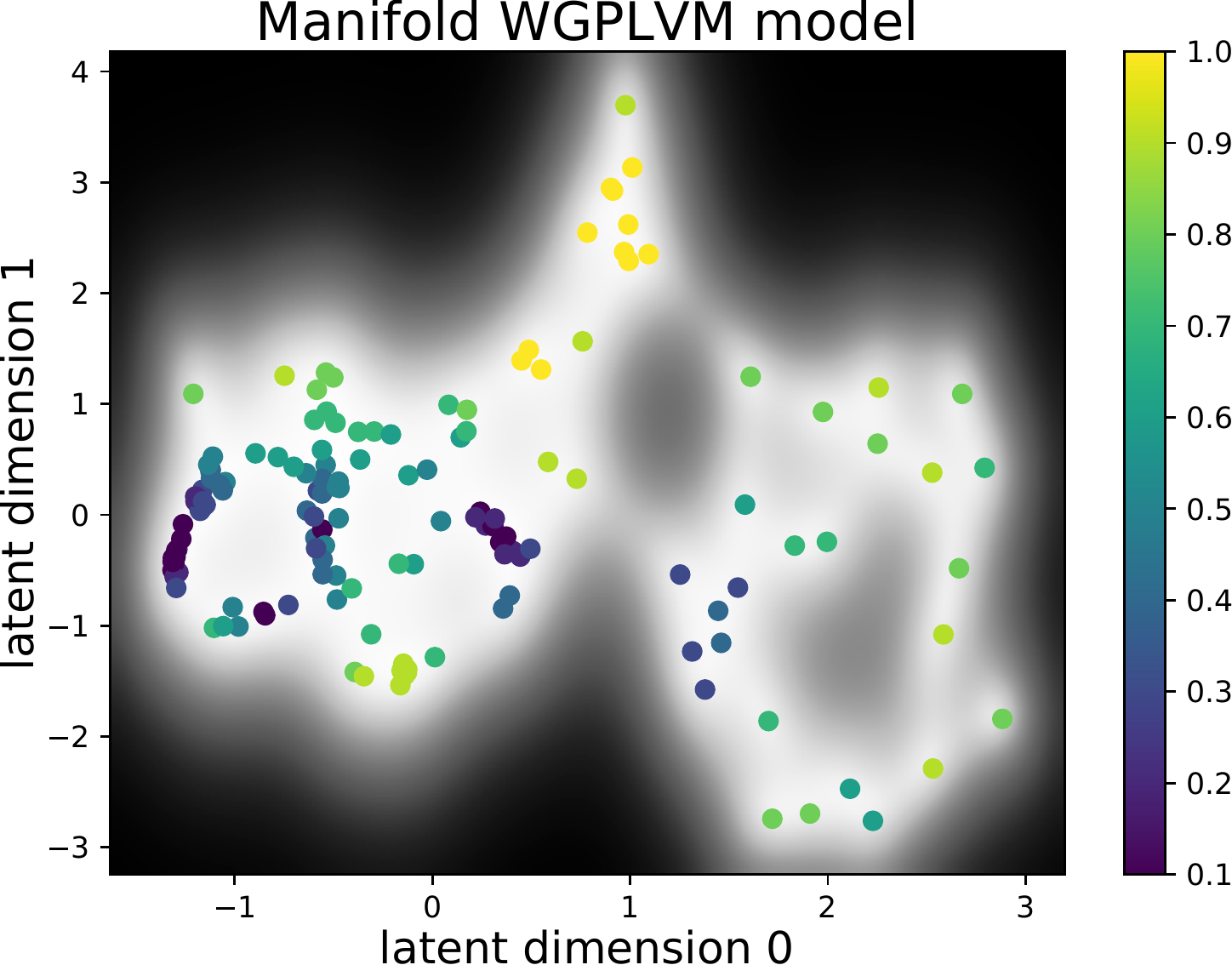} \hfil \includegraphics[width=0.45\linewidth]{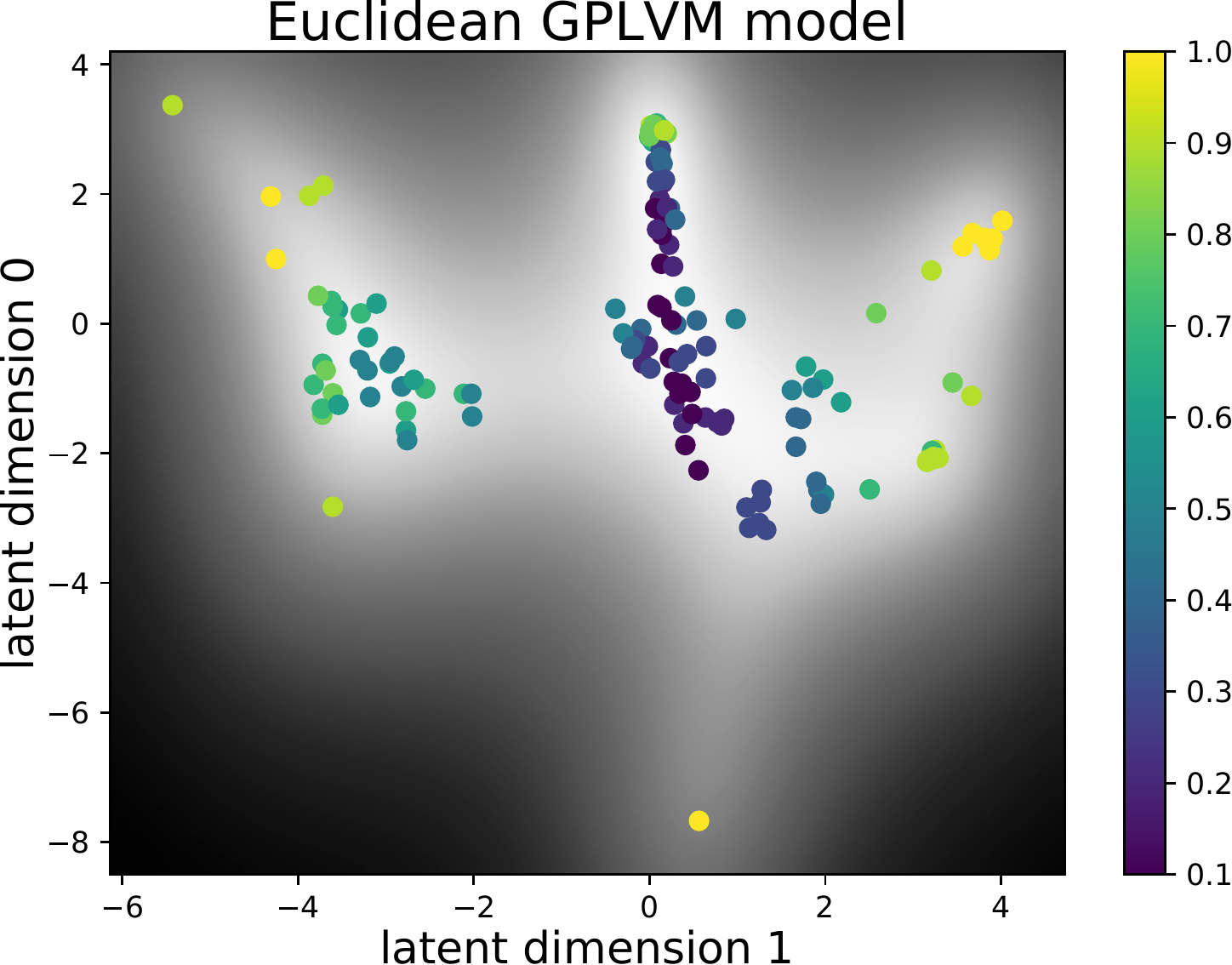}
\caption{The latent spaces for the diffusion-tensor dataset learned using the WGPLVM and GPLVM models. The colors indicate the FA of the given tensor.}
\label{fig:latent_dti}
\end{figure}

The \emph{fractional anisotropy} (FA) of a $3 \times 3$ SPD matrix is a shape descriptor taking values between $0$ and $1$, where an FA of $0$ corresponds to a round tensor, and an FA near $0$ corresponds to a very thin one. Given the eigenvalues $\lambda_1, \lambda_2, \lambda_3$ for an SPD matrix, its FA is defined as 
\[
\sqrt{\frac{3}{2}} \frac{\sqrt{(\lambda_1 - \hat{\lambda})^2 + (\lambda_2 - \hat{\lambda})^2 + (\lambda_3 - \hat{\lambda})^2}}{\sqrt{\lambda_1^2 + \lambda_2^2 + \lambda_3^2}},
\]
where $\hat{\lambda}$ is the mean of the eigenvalues. In the latent space shown in Fig.~\ref{fig:latent_dti}, the latent variables are colored according to the FA of their associated tensor, and we see that both models provide a smooth transition between different FA values.

\begin{figure}[h!]
\includegraphics[width=0.45\linewidth]{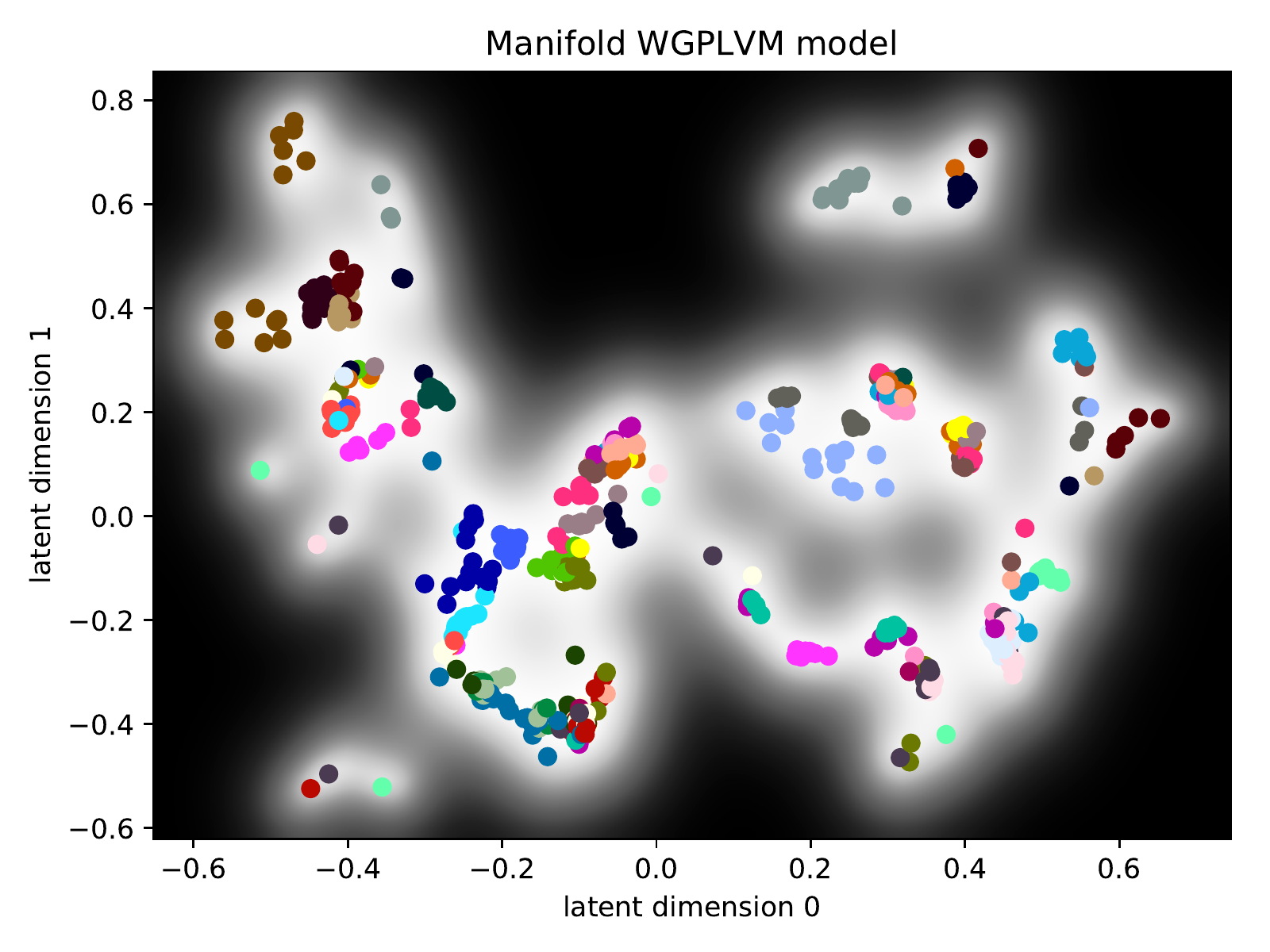} \hfil \includegraphics[width=0.45\linewidth]{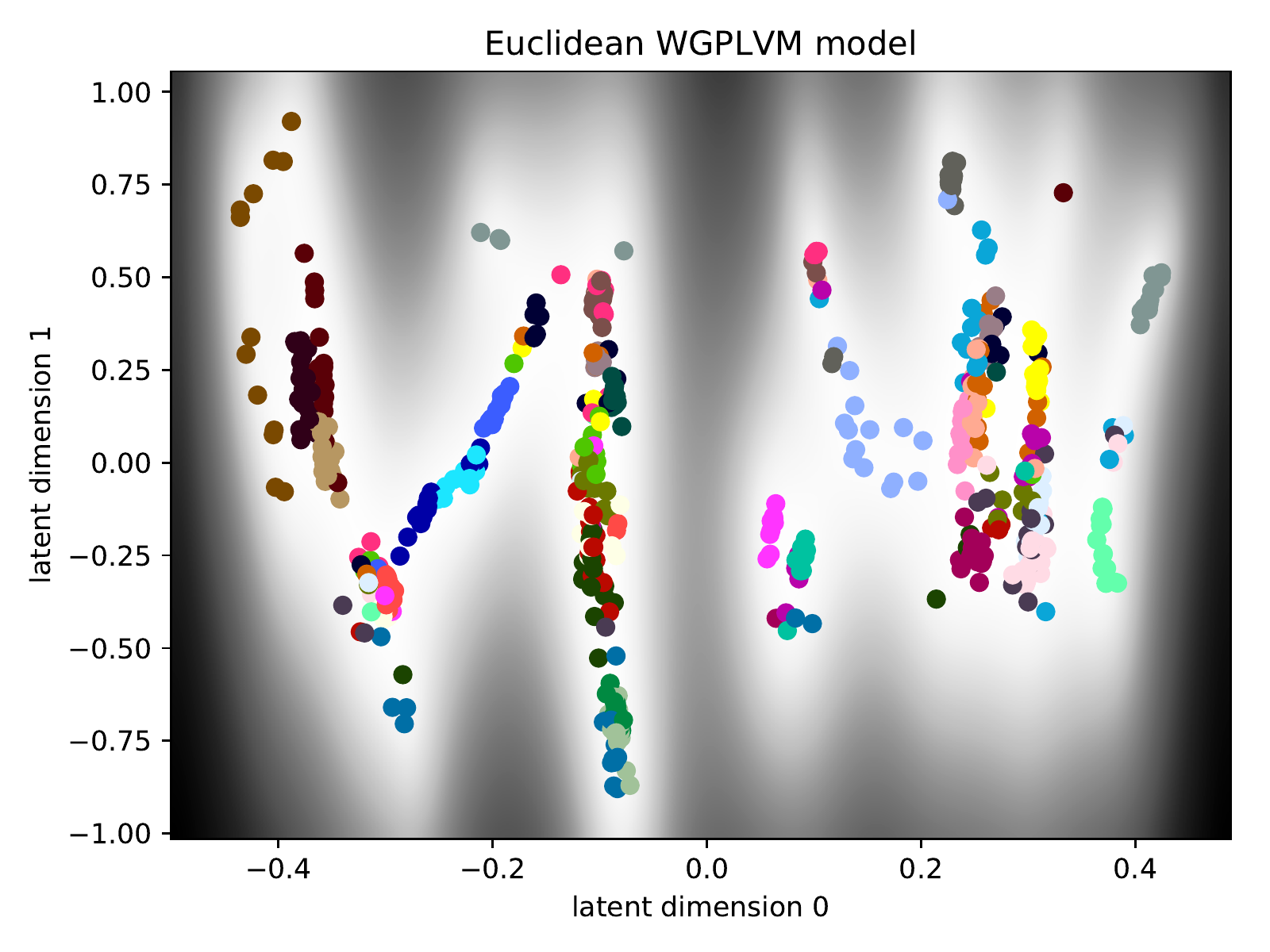}
\caption{The latent spaces for the diatom dataset learned using the WGPLVM and GPLVM models. The colors indicate the species of the diatom corresponding to the latent variable, see Fig. \ref{diatoms}.}
\label{fig:latent_diatom}
\end{figure}

\begin{figure*}[h!]
\centering
\includegraphics[width = .9\textwidth]{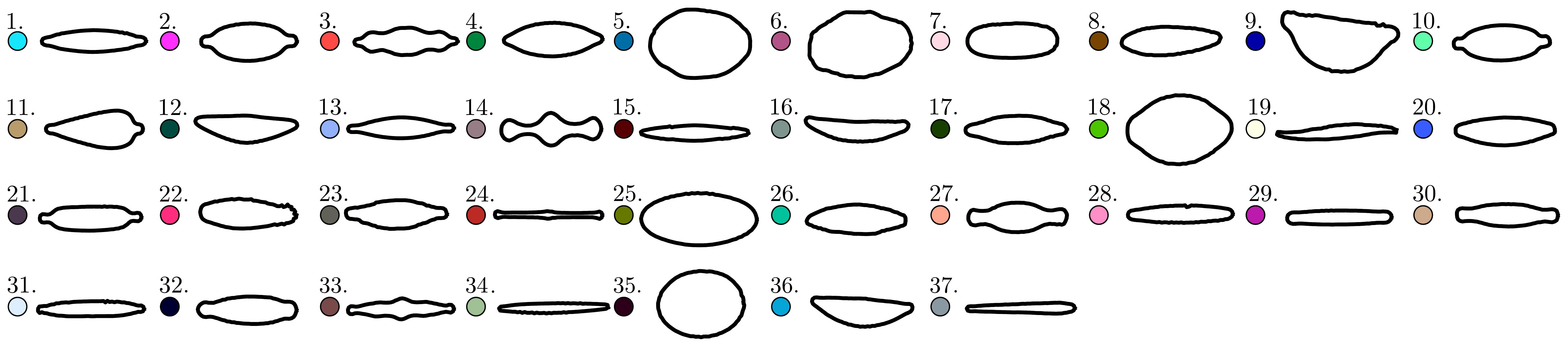}
\caption{Representatives of each of the $37$ diatom classes with corresponding class colors used in Fig. \ref{fig:latent_diatom}. Note that variation inside of each class can be considerable.}
\label{diatoms}
\end{figure*}

The latent space visualization of the diatom dataset is found in Fig.~\ref{fig:latent_diatom}; here the latent variables are colored by the species of the corresponding diatom, see Fig.~\ref{diatoms} for a visualization of species representatives.

\begin{figure*}[h!]
\centering
\includegraphics[width = 1\textwidth]{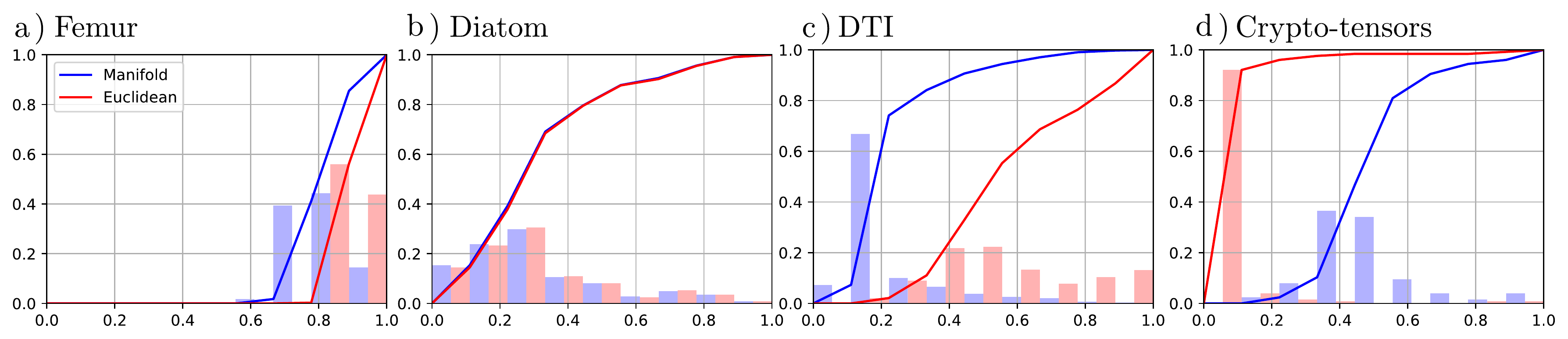}
\caption{Distributions of distances between the data points and the population means. The bar plots indicate the density of data points that lie x-fraction of the maximum distance away from the mean. The corresponding continuous curves represent the cumulative distributions.}
\label{geomcomp}
\end{figure*}
\subsection{Comparing the Geometries}
In this section, we compare the geometries in Euclidean and Riemannian cases. The aim is to try and understand, when the performance is improved. We do this by visualizing the distribution of data point distances to the corresponding population means, the distances and means computed according to the corresponding metrics. 

As can be seen in Fig. \ref{geomcomp}, in the femur ($2$-sphere) and diatom (Kendall's shape space) cases, the distributions look very similar. In fact, in the diatom case, they are essentially the same. The Kendall's shape space forms a quotient manifold of the sphere, which in this case is high dimensional ($d = 180$). In such high dimension, escaping the manifold becomes increasingly more difficult (most of the volume of the sphere is close to the boundary), and thus both the metrics are essentially the same. This might explain, why the WGPLVM did not improve notably on the GPLVM.

In the crypto-tensor experiment, the distribution implies the presence of extreme outliers under the Euclidean metric. The Log-Euclidean metric, on the other hand, transforms the metric scale, evening out the distribution. This could very well explain, why we see large improvement with the WGPLVM compared to the GPLVM.

In the DTI experiment, the distribution of Euclidean distances looks more even. This might imply, that in this occasion, the Euclidean distance is better at capturing the trend of the data. However, the improved uncertainty estimates of the WGPLVM could be explained, as the Euclidean models are not confined to $SPD(n)$. Therefore, the distributions do not follow the conic shape of $SPD(n)$. 

\end{document}